\documentclass[journal,comsoc]{IEEEtran}
\usepackage[T1]{fontenc}

\ifCLASSINFOpdf
\else
\fi
\usepackage{graphicx}
\usepackage{caption}
\usepackage{subcaption}
\usepackage{enumitem}
\usepackage{amsfonts}
\usepackage{bbm}
\usepackage{footnote}
\usepackage{colortbl}
\usepackage{amsmath}
\usepackage{empheq}
\usepackage{xcolor}
\usepackage{makecell}
\usepackage{enumitem}
\usepackage{stmaryrd}
\usepackage{array}
\newcolumntype{C}[1]{>{\centering\arraybackslash}m{#1}}
\definecolor{myRed}{rgb}{1.0,0.5,0.5}
\definecolor{myBlue}{rgb}{0.5,0.5,1.0}

\usepackage[cmintegrals]{newtxmath}
\usepackage{lipsum}
\usepackage{xpatch}

\makeatletter
\newinsert\copyrightnote@ins
\count\copyrightnote@ins=1000
\dimen\copyrightnote@ins=8in
\newcommand*\copyrightnote@hook
  {%
    \unvbox\copyrightnote@ins
    \global\let\@makecol\copyrightnote@makecol
  }
\let\copyrightnote@AtBeginDocument\AtBeginDocument
\AtBeginDocument{\let\copyrightnote@AtBeginDocument\@firstofone}
\newcommand*\copyrightnote@firstuse
  {%
    \gdef\copyrightnote@firstuse
      {\global\skip\copyrightnote@ins=\bigskipamount\gdef\copyrightnote@firstuse{}}%
    \global\let\copyrightnote@makecol\@makecol
    \xpatchcmd\@makecol{\unvbox\footins}{\unvbox\footins\copyrightnote@hook}
      {}{\GenericError{}{patching @makecol failed}{}{}}
    \copyrightnote@AtBeginDocument
      {%
        \ifvoid\footins
          \insert\footins{}%
        \else
          \global\skip\copyrightnote@ins=\bigskipamount
        \fi
      }%
  }
\newcommand\copyrightnote[1]
  {%
    \copyrightnote@firstuse
    \copyrightnote@AtBeginDocument
      {%
        \insert\copyrightnote@ins
          {%
            \reset@font\footnotesize
            \interlinepenalty\interfootnotelinepenalty
            \splittopskip\footnotesep
            \splitmaxdepth\dp\strutbox
            \floatingpenalty\@MM
            \hsize\columnwidth
            \@parboxrestore
            \color@begingroup
            \parindent1em
            \noindent
            \hskip1.8em
            \ignorespaces #1\@finalstrut\strutbox
            \color@endgroup
          }%
      }%
  }

\begin{document}

\title{Human Trajectory Forecasting in Crowds:\\ A Deep Learning Perspective}

%

%
%
%

\author{Parth~Kothari, 
        Sven~Kreiss, 
        and~Alexandre~Alahi}%
\maketitle
\copyrightnote{This work has been submitted to the IEEE for possible publication. Copyright may be transferred without notice, after which this version may no longer be accessible.}

\begin{abstract}
Since the past few decades, human trajectory forecasting has been a field of active research owing to its numerous real-world applications: evacuation situation analysis, deployment of intelligent transport systems, traffic operations, to name a few. Early works handcrafted this representation based on domain knowledge. However, social interactions in crowded environments are not only diverse but often subtle. Recently, deep learning methods have outperformed their handcrafted counterparts, as they learned about human-human interactions in a more generic data-driven fashion. In this work, we present an in-depth analysis of existing deep learning-based methods for modelling social interactions. We propose two knowledge-based data-driven methods to effectively capture these social interactions. To objectively compare the performance of these interaction-based forecasting models, we develop a large scale interaction-centric benchmark \textit{TrajNet++}, a significant yet missing component in the field of human trajectory forecasting. We propose novel performance metrics that evaluate the ability of a model to output socially acceptable trajectories. Experiments on TrajNet++ validate the need for our proposed metrics, and our method outperforms competitive baselines on both real-world and synthetic datasets.
\end{abstract}



%
\IEEEpeerreviewmaketitle

\section{Introduction}
\label{sec1}

Humans possess the natural ability to navigate in social environments. In other words, we have understood the social etiquettes of human motion like respecting personal space, yielding right-of-way, avoid walking through people belonging to the same group. Our social interactions lead to various complex pattern-formation phenomena in crowds, for instance, the emergence of lanes of pedestrians with uniform walking direction, oscillations of the pedestrian flow at bottlenecks. The ability to model social interactions and thereby forecast crowd dynamics in real-world environments is extremely valuable for a wide range of applications: infrastructure design \cite{Jiang1999SimPedSP, Lerner2007CrowdsBE, Bitgood2006AnAO}, traffic operations \cite{Horni2016TheMT}, crowd abnormality detection systems \cite{Mehran2009AbnormalCB}, evacuation situation analysis \cite{Helbing2002SimulationOP, Helbing2000SimulatingDF, Zheng2009ModelingCE, Moussad2011HowSR, Dong2020StateoftheArtPA}, deployment of intelligent transport systems \cite{WaymoSafety, UberSafety, Chen2019CrowdRobotIC, Rasouli2020AutonomousVT} and recently helping in the broad quest of building a digital twin of our built environment. However, modelling social interactions is an extremely challenging task as there exists no fixed set of rules which govern human motion. A task closely related to learning human social interactions is forecasting the movement of the surrounding people, which conform to common social norms. We refer to this task of forecasting the human motion as \textit{human trajectory forecasting.}

\begin{figure}
    \centering
    \includegraphics[width=0.35\textwidth]{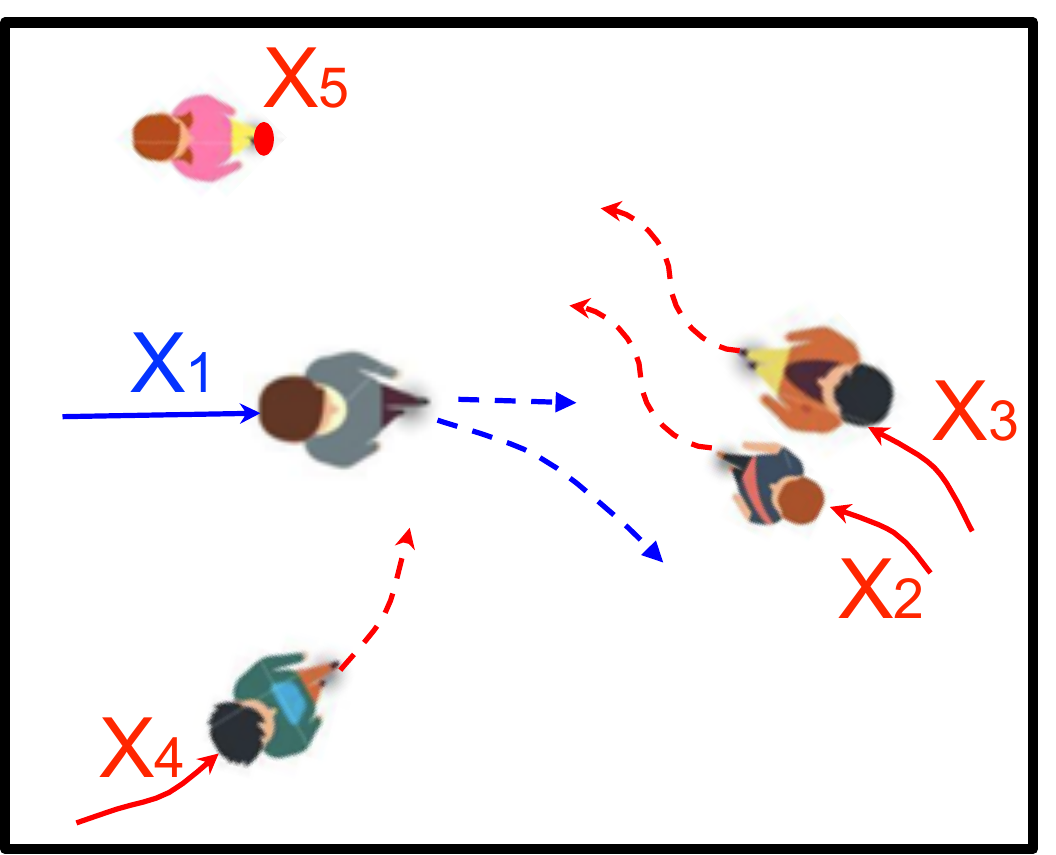}
    \caption{Human trajectory forecasting is the task of forecasting the future trajectories (dashed) of all humans which conform to the social norms, given the past observed scene (solid). The presence of social interactions distinguish human trajectory forecasting from other sequence modelling tasks: the primary pedestrian (X1) deviates from his direction of motion to avoid a collision, by forecasting the trajectory of the child (X2).}
    \label{fig:PullFigure}
\end{figure}

Before formally defining human trajectory forecasting, we introduce the notion of \textit{Trajectory} and \textit{Scene}. We define a \textit{Trajectory} as the time-profile of pedestrian motion states. Generally, these states are the position and velocity of a human. However, we can consider more complex states like body pose, to glean more information about a person's movement. We define a \textit{Scene} as a collection of trajectories of multiple humans interacting in a social setting. A scene may also comprise physical objects and non-navigable areas that affect the human trajectories, \textit{e.g.,} walls, doors, and elevators. Also, when required, we refer to a particular pedestrian of interest in the scene as the \textit{Primary pedestrian.} We define human trajectory forecasting as follows:\\[0.3cm]
\textit{Given the past trajectories of all humans in a scene, forecast the future trajectories which conform to the social norms.}\\[0.2cm]
Human trajectory forecasting is primarily a sequence modelling task. The typical challenges for a sequence modelling task are (1) encoding observation sequence: we need to learn to model the long-term dependency in the past trajectory effectively, (2) multimodality: given the history of a scene, multiple futures (predictions) are plausible. In addition to this, for human trajectory forecasting, there exist two crucial challenges that differentiate it from other sequence prediction tasks such as language modelling, weather forecasting, and stock market forecasting (see Fig~\ref{fig:PullFigure}):
\begin{itemize}[leftmargin=*]
\itemsep0em
    \item \textbf{Presence of social interactions}: the trajectory of a person is affected by the motion of the other people in his/her surroundings. Modelling how the observation of one sequence affects the forecast of another sequence is an essential requirement for a good human trajectory forecasting model.
    \item \textbf{Physically acceptable outputs}: a good human trajectory forecasting model should provide physically acceptable outputs, for instance, the model prediction should not undergo collisions. Quantifying the physical feasibility of a model prediction is crucial for safety-critical applications.  
\end{itemize}

Our objective is to encode the observed scene into a representation that captures all information necessary to forecast human motion. To focus on learning the social interactions that affect social motion, we assume that there do not exist any physical constraints in our scenes. The future trajectory can also be affected by the long-term goal of the human, which cannot always be observed or inferred. We therefore focus on \textit{short-term} human trajectory forecasting (next 5 secs).

Following the success of Social LSTM \cite{Alahi2016SocialLH}, a variety of neural networks (NN) based modules that model social interactions have been proposed in literature. In this work, we explicitly focus on the design of these interaction modules and not the entire forecasting model. The challenge in designing these interaction modules lies in handling a variable number of neighbours and modelling how they collectively influence one's future trajectory. We present a high-level pipeline encompassing most of the existing designs of interaction modules. Based on our taxonomy, we propose two novel modules which incorporate domain knowledge into the NN-based pipeline. As a consequence, these modules are better equipped to learn social etiquettes like collision avoidance and leader-follower. 
A long-standing question in NN-based trajectory forecasting models is to explore techniques that help to explain the model decisions. In this work, we propose to utilize Layer-wise Relevance Propagation (LRP) \cite{Bach2015OnPE} to explain the decisions of our trajectory forecasting models. To the best of our knowledge, this is the first work that applies LRP in a \textit{regression} setting, to infer inter-sequence (neighbours) effects on the model output.  

To demonstrate the efficacy of a trajectory forecasting model, one needs to have the means to objectively compare with other forecasting baselines on good quality datasets. However, current methods have been evaluated on different subsets of available data without a proper sampling of scenes in which social interactions occur. As our final contribution, we introduce \textbf{TrajNet++}, a large scale interaction-centric trajectory forecasting benchmark comprising explicit agent-agent scenarios. Our benchmark provides proper indexing of trajectories by defining a hierarchy of trajectory categorization. In addition, we provide an extensive evaluation system to test the gathered methods for a fair comparison. In our evaluation, we go beyond the standard distance-based metrics and introduce novel metrics that measure the capability of a model to emulate pedestrian behavior in crowds. We demonstrate the efficacy of our proposed domain-knowledge based baselines on TrajNet++, in comparison to various interaction encoder designs. Furthermore, we illustrate how the decisions of our proposed model architecture can be explained using LRP in real-world scenarios. 


To summarize, our main contributions are as follows:
\begin{enumerate}
    \item We provide an in-depth analysis of existing designs of NN-based interaction encoders along with their source code. We explain the decision-making of trajectory forecasting models by extending layer-wise relevance propagation to the regression setting of trajectory forecasting.
    \item We propose two simple yet novel methods driven by domain knowledge for capturing social interactions.  
    \item We present TrajNet++, a large scale interaction-centric trajectory forecasting benchmark with novel evaluation metrics that quantify the \textit{physical feasibility} of a model. 
\end{enumerate}




%

\section{Related Work}
\label{sec2}

Finding the ideal representation to encode human social interactions in crowded environments is an extremely challenging task. Social interactions are not only diverse but often subtle. In this work, we consider microscopic models of pedestrian crowds, where collective phenomena emerge from the complex interactions between many individuals (self-organizing effects). Current human trajectory forecasting works can be categorized into learning human-human (social) interactions or human-space (physical) interactions or both. Our work is focused on deep learning based models that capture social interactions. In this section, we review the work done for modelling the human-human interactions to obtain the social representation. 


With a specific focus on pedestrian path forecasting problem, Helbing and Molnar \cite{SocialForce} presented a force-based motion model with attractive forces (towards the goal of a person and towards his/her group) and repulsive forces (away from people not belonging to a person's group and physical obstacles), called Social Force model, which captures the social and physical interactions. Their seminal work displays competitive results even on modern pedestrian datasets and has been extended for improved trajectory forecasting \cite{Elfring2013LearningIF, Rudenko2018HumanMP, Rudenko2018JointLP, Yu2020ImprovedOI}, tracking \cite{Luber2010PeopleTW, Pellegrini2010ImprovingDA, LealTaix2014LearningAI} and activity forecasting \cite{Choi2012AUF, Choi2014UnderstandingCA}. Burstedde \textit{et al.} \cite{Burstedde2001SimulationOP} utilize the cellular automaton model, another type of microscopic model, for predicted pedestrian motion. In their model, the environment is divided into uniformly distributed grids and each pedestrian has a matrix of preference to determine the transition to neighbouring cells. The matrix of preference is determined by the pedestrian's own intent along with the locations of surrounding agents. Similar to social force, the cellular automaton model has been extended over the years for improved trajectory forecasting \cite{Vizzari2015AnAP}. Another prominent model for simulating human motion is Reciprocal Velocity Obstacles (RVO) \cite{Berg2008ReciprocalVO}, which guarantees safe and oscillation-free motion, assuming that each agent follows identical collision avoidance reasoning. Social interaction modelling has been approached from different perspectives such as Discrete Choice framework \cite{DiscreteChoice}, continuum dynamics \cite{Treuille2006ContinuumC} and Gaussian processes \cite{Keat2007ModellingSP, Kim2011GaussianPR, Mnguez2019PedestrianPP}.  Robicquet \textit{et al.} \cite{Robicquet2016LearningSE} defined social sensitivity to characterize human motion into different navigation styles. Alahi \textit{et al.} \cite{Alahi2014SociallyAwareLC} defined Social Affinity Maps to link broken or unobserved trajectories to forecast pedestrian destinations. Yi \textit{et al.} \cite{Yi2015UnderstandingPB} exploited crowd grouping as a cue to better forecast trajectories. However, all these methods use handcrafted functions based on relative distances and specific rules to model interactions. These functions impose not only strong priors but also have limited capacity when modelling complex interactions. In recent times, methods based on neural networks (NNs) that infer interactions in a data-driven fashion have been shown to outperform the works mentioned above.    


Inspired by the application of recurrent neural networks (RNNs) in diverse sequence prediction tasks \cite{Graves2013SpeechRW, Meng2014NeuralMT, Vinyals2015ShowAT, Cao2015LookAT}, Alahi et al. \cite{Alahi2016SocialLH} proposed Social LSTM, the first NN-based model for human trajectory forecasting. Social LSTM is an LSTM \cite{Hochreiter1997LongSM} network with a novel social pooling layer to capture social interactions of nearby pedestrians. RNNs incorporating social interactions allow anticipating interactions that can occur in a more distant future. The social pooling module has been extended to incorporate physical space context \cite{Varshneya2017HumanTP, Lee2017DESIREDF, Bartoli2018ContextAwareTP, Xue2018SSLSTMAH, Zhao2019MultiAgentTF, Lisotto2019SocialAS} and various other designs of NN-based interaction module have been proposed \cite{Pfeiffer2017ADM, Shi2019PedestrianTP, Bisagno2018GroupLG, Gupta2018SocialGS, Zhang2019SRLSTMSR, Zhu2019StarNetPT, Ivanovic2018TheTP, Liang2019PeekingIT, Tordeux2019PredictionOP, Ma2016AnAI, Hasan2018MXLSTMMT, Hasan2019ForecastingPT, Salzmann2020TrajectronMG}. Pfieffer \textit{et al.} \cite{Pfeiffer2017ADM} proposed an angular pooling grid for efficient computation. Shi \textit{et al.} \cite{Shi2019PedestrianTP} proposed an elliptical pooling grid placed along the direction of movement of the pedestrian with more focus on the pedestrians in the front. Bisagno \textit{et al.} \cite{Bisagno2018GroupLG} proposed to consider only pedestrians not belonging to the same group during social pooling. While modelling social interactions, Hasan \textit{et al.} \cite{Hasan2018MXLSTMMT, Hasan2019ForecastingPT} based on domain knowledge, only consider the pedestrians in the visual frustum of attention \cite{Hasan2018SeeingIB}.  Gupta \textit{et al.} \cite{Gupta2018SocialGS} propose to encode neighbourhood information through the use of a permutation-invariant (symmetric) max-pooling function. Zhang \textit{et al.} \cite{Zhang2019SRLSTMSR} proposed to refine the state of the LSTM cell using message passing algorithms. Zhu \textit{et al.} \cite{Zhu2019StarNetPT} proposed a novel star topology to model interactions. The center hub maintains information of the entire scene which each pedestrian can query. Ivanovic \textit{et al.} \cite{Ivanovic2018TheTP} and Salzmann \textit{et al.} \cite{Salzmann2020TrajectronMG} proposed to sum-pool the neighbour states and pass it through an LSTM-based encoder to obtain the interaction vector. Liang \textit{et al.} \cite{Liang2019PeekingIT} proposed to utilize geometric relations obtained from the spatial distance between pedestrians, to derive the interaction representation. \cite{Tordeux2019PredictionOP, Ma2016AnAI} propose to input the relative position and relative velocity of $k$ nearest neighbours directly to an MLP to obtain the interaction vector. Many works \cite{Xu2018EncodingCI, Fernando2018SoftH, Li2020SocialWaGDATIT, Sadeghian2018SoPhieAA, Kosaraju2019SocialBiGATMT, Amirian2019SocialWL, Fernando2018GDGANGA, Haddad2019SituationAwarePT, Huang2019STGATMS, Mohamed2020SocialSTGCNNAS, Li2020EvolveGraphHM, Sun2020RecursiveSB, Yu2020SpatioTemporalGT, Mangalam2020ItIN, Tao2020DynamicAS} propose interaction module designs based on attention mechanisms \cite{Vaswani2017AttentionIA, Bahdanau2014NeuralMT} to identify the neighbours which affect the trajectory of the person of interest. The attention weights are either learned or handcrafted based on domain knowledge (\textit{e.g.}, euclidean distance). For an extensive survey of all human forecasting methods capturing both social and physical interactions, one can refer to Rudenko \textit{et al.} \cite{Rudenko2019HumanMT}.

\section{Problem Statement}

Our objective is to forecast the future trajectories of all the pedestrians present in a scene. The network takes as input the trajectories of all the people in a scene denoted by $\mathbf{X} = \{X_{1}, X_{2}, \hdots, X_{n}\}$ and our task is to forecast the corresponding future trajectories $\mathbf{Y} = \{Y_{1}, Y_{2}, \hdots, Y_{n}\}$. The position and velocity of pedestrian $i$ at time-step $t$ is denoted by $\mathbf{x}^{t}_{i} = (x^{t}_{i} , y^{t}_{i})$ and $\mathbf{v}^{t}_{i}$ respectively. We receive the positions of all pedestrians at time-steps $t = 1, \hdots, T_{obs}$ and want to forecast the future positions from time-steps $t = T_{obs + 1}$ to $T_{pred}.$ We denote our predictions using $\mathbf{\hat{Y}}$.

At time-step $t$, we denote the state of pedestrian $i$ by $\mathbf{s}^{t}_{i}$. The state can refer to different attributes of the person, \textit{e.g.}, the position concatenated with velocity ($\mathbf{s}^{t}_{i} = [\mathbf{x}^{t}_{i}, \mathbf{v}^{t}_{i}]$). The problem statement can be extended to take as input more attributes at each time-step, \textit{e.g.}, the body pose, as well as predicting $k$ most-likely future trajectories. 

\begin{figure}[h!]
    \centering
    \includegraphics[width=0.48\textwidth]{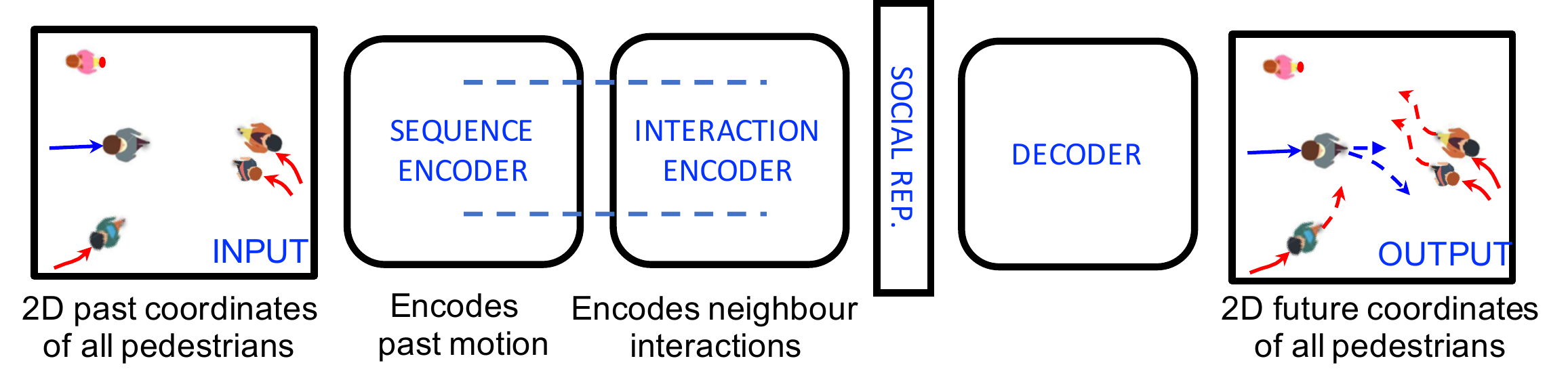}
    \caption{A data-driven pipeline for human trajectory forecasting. 
    We focus on the design choices for the interaction module.}
    \label{fig:global_pipeline}
\end{figure}

\section{Method}
\label{sec3}




A global data-driven pipeline for forecasting human motion is illustrated in Fig~\ref{fig:global_pipeline}. It comprises of the motion encoding module, the interaction module and the decoder module. On a high level, the motion encoding module is responsible for encoding the past motion of pedestrians. The interaction module learns to capture the social interactions between pedestrians. The motion encoding module and the interaction module are not necessarily mutually exclusive. The output of the interaction module is the social representation of the scene. The social representation is passed to the decoder module to predict a single trajectory or a trajectory distribution depending on the decoder architecture. Since our benchmark TrajNet++ focuses on interaction-centric scenes, in this work, we focus on investigating the design choices for the interaction module. 

\begin{figure*}[t]
\begin{subfigure}[h]{0.30\textwidth}
\includegraphics[width=\textwidth]{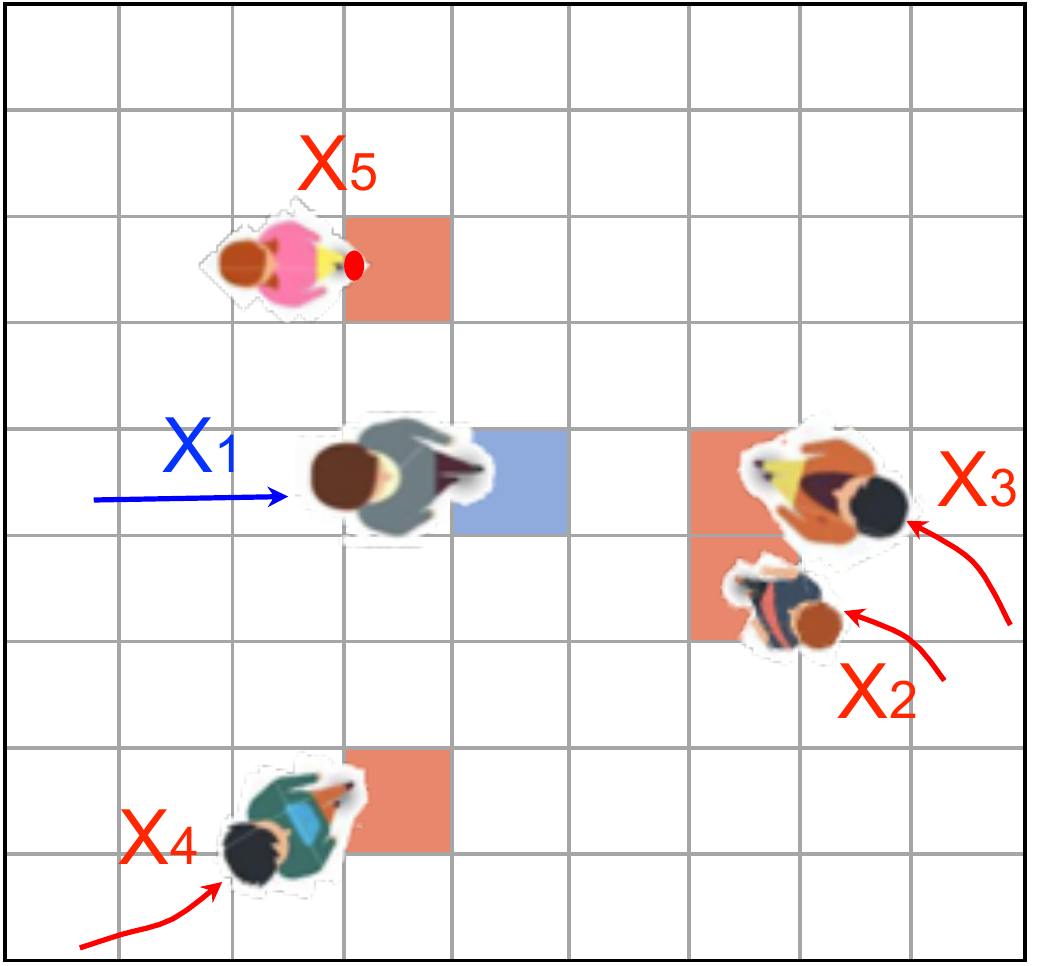}
\caption{Occupancy pooling \cite{Alahi2016SocialLH}}
\label{fig:occ}
\end{subfigure}
\begin{subfigure}[h]{0.30\textwidth}
\includegraphics[width=\textwidth]{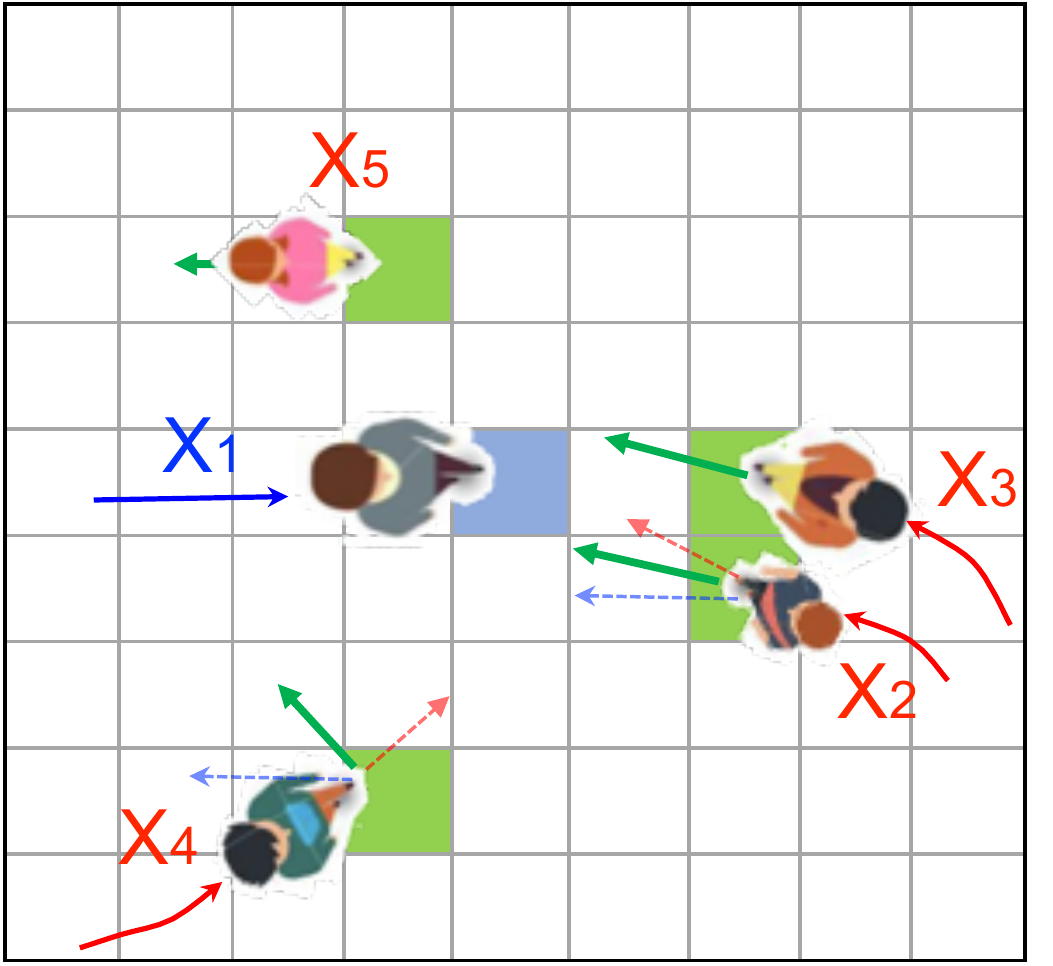}
\caption{Directional pooling [Ours]}
\label{fig:dir}
\end{subfigure}
\begin{subfigure}[h]{0.38\textwidth}
\includegraphics[width=\textwidth]{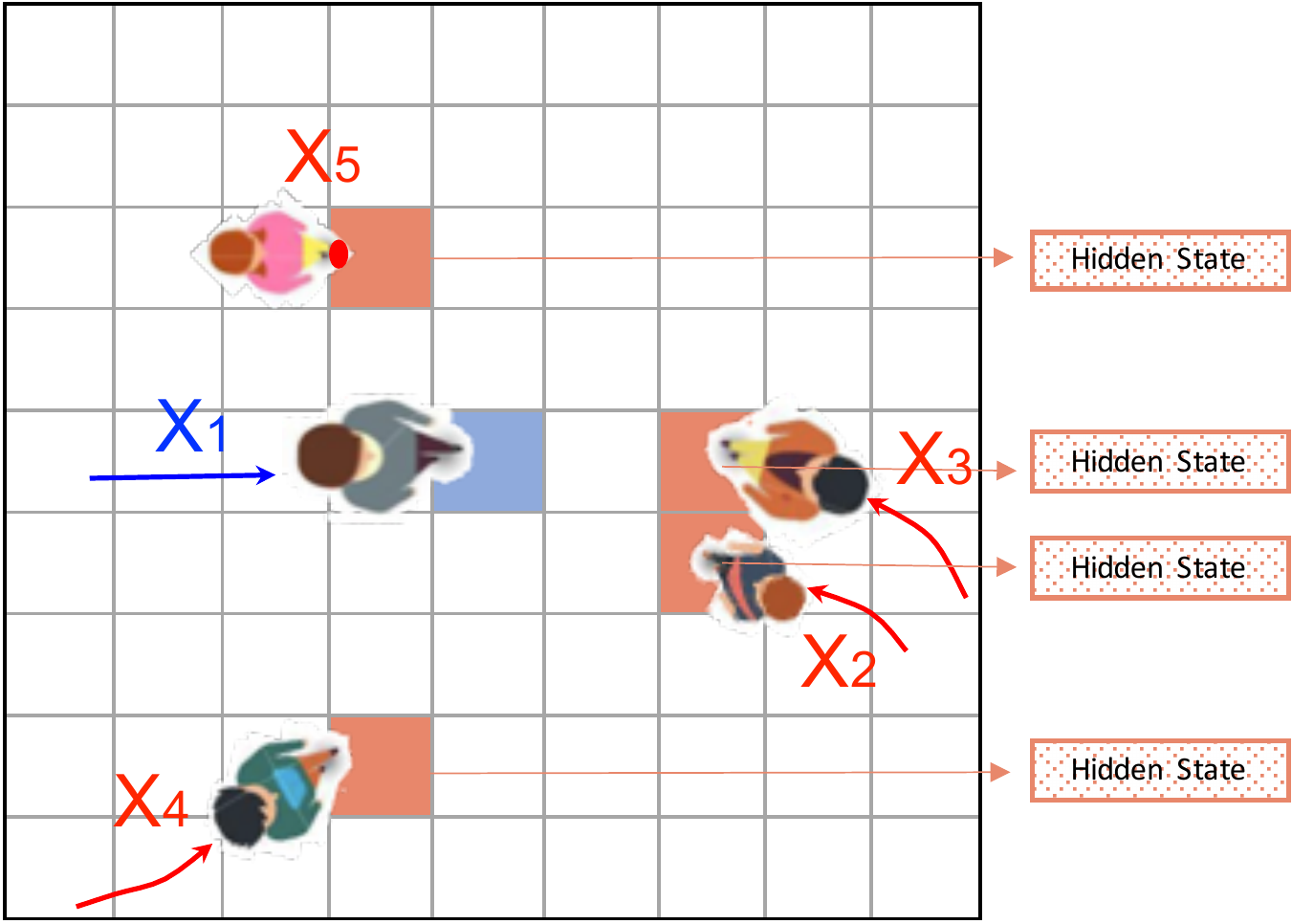}
\caption{Social pooling \cite{Alahi2016SocialLH}}
\label{fig:soc}
\end{subfigure}\\
\caption{Illustration of the grid-based interaction encoding modules. (a) Occupancy pooling: each cell indicates the presence of a neighbour (b) Our proposed directional pooling: each cell contains the relative velocity of the neighbour with respect to the primary pedestrian. (c) Social pooling: each cell contains the LSTM hidden-state of the neighbour. The constructed grid tensors are passed through an MLP-based neural network to obtain the interaction vector.}
\label{fig:grid_pool}
\end{figure*}

\subsection{Interaction Module}




Humans have the capability to navigate with ease in complex, crowded environments by following unspoken social rules, which result in social interactions. In recent years, these social interactions are captured effectively by designing novel interaction modules. In this section, we broadly categorize the different data-driven interaction encoders studied in literature, based on their underlying components. We show how most of these designs fall within our categorization. Following this, in the experimental section we empirically analyze the effectiveness of each of these components and provide recommendations for designing improved interaction modules. The existing designs can be broadly categorized into (1) \textbf{Grid based} and (2) \textbf{Non-Grid based}. We now discuss in detail the different components of these interaction encoders.

\subsubsection{\textbf{Grid Based Interaction Models}}
In grid-based models, the interaction module takes as input a local grid constructed around the pedestrian of interest. Each cell within the grid represents a particular spatial position relative to the pedestrian of interest. The design of grid-based models largely differ based on neighbour input state representation.

\textbf{Neighbour input state:} Consider an $N_o \times N_o$ grid around the primary pedestrian, where each cell contains information about neighbours located in that corresponding position. Existing designs provides the information of the neighbours in two main forms: (a) \textit{Occupancy Pooling} \cite{Alahi2016SocialLH, Bartoli2018ContextAwareTP} where each cell in the grid indicates the presence of a neighbour (see Fig~\ref{fig:occ}) (b) \textit{Social Pooling} \cite{Alahi2016SocialLH, Bartoli2018ContextAwareTP, Zhao2019MultiAgentTF, Bisagno2018GroupLG, Varshneya2017HumanTP, Lisotto2019SocialAS, Lee2017DESIREDF} where each cell contains the entire past history of the neighbour, represented by, \textit{e.g.}, the LSTM hidden state of the neighbours (see Fig~\ref{fig:soc}). The obtained grid is embedded using an MLP to get the interaction vector $p^t_{i}$.


\textbf{Directional Pooling} \\
In this work, based on our domain knowledge, we propose to take as input the relative velocity of each neighbour in the corresponding grid cell. When humans navigate in crowded environments, in addition to relative positions of the neighbours, they naturally tend to focus on the neighbours’ velocities. For the same positional configuration, the relative velocities of neighbours lead to the concepts of leader-follower and collision avoidance \textit{i.e.,} one exhibits leader-follower and accelerates when the neighbour is in front and walking along the same direction, while the same positional configuration leads to deceleration when the neighbour moves in the opposite direction. Having access to relative velocities can therefore significantly improve model performance in preventing collisions.


Furthermore, due to the complex nature of real-world movements combined with the possibility of noisy measurements, the current design of social pooling can sometimes fail to learn the important notion of preventing collisions. One reason lies in the fact that the models are trained to minimize the displacement errors \cite{Alahi2016SocialLH, Kosaraju2019SocialBiGATMT} and not collisions. The models are expected to learn the notion of collision avoidance implicitly. By focusing explicitly on relative velocity configurations, we can obtain more domain-knowledge driven control over the design of the interaction encoder. When the model explicitly focuses only on relative velocity configuration (rather than abstract hidden-state configurations), which is sufficient to learn concepts of leader-follower and collision avoidance, the resulting simple design has the potential to output safer predictions. Furthermore, directional pooling design are computationally faster to deploy in real-time scenarios due to the reduced size of input ($N \times N \times 2$ in comparison to $N \times N \times H_{dim}$ where $H_{dim}$ is the dimension of the hidden-state).

One might additionally argue to only consider the neighbours in front of the primary pedestrian as proposed in \cite{Hasan2018SeeingIB}. We will demonstrate in the experimental section that the directional pooling implicitly learns this notion of only focusing on the neighbours in the field-of-view of the primary pedestrian.


\begin{figure*}[ht]
\centering
\begin{subfigure}[h]{0.99\textwidth}
\includegraphics[width=0.99\textwidth]{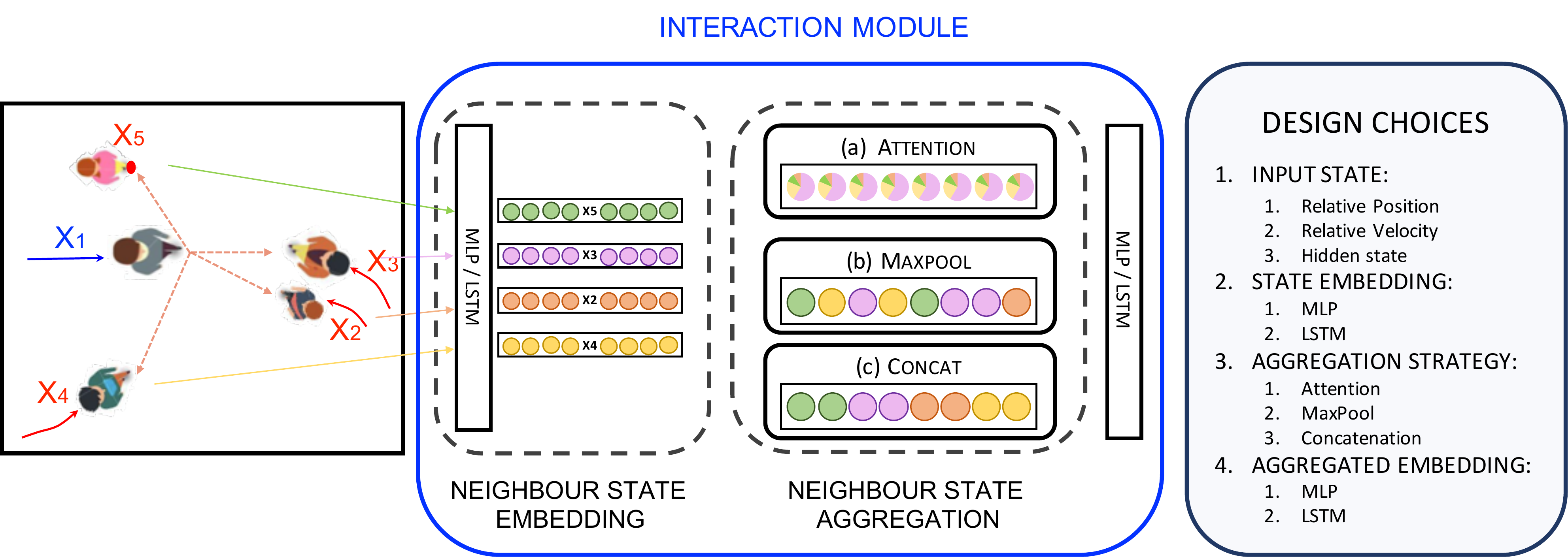}
\end{subfigure}
\centering 
\caption{Illustration of the non-grid based encoding modules to obtain the interaction vector. The challenge lies in handling a variable number of neighbours and aggregating their state information to construct the interaction vector (a) Neighbour information is aggregated via attention mechanism (b) Neighbour information is aggregated utilizing a symmetric function
(c) Neighbour information is aggregated via concatenation.}
\label{fig:mix_grid_pool}
\end{figure*}

\subsubsection{\textbf{Non-Grid Based Interaction Models}}
Non-grid based modules, as the name suggests, capture the social interactions in a grid-free manner. The challenge in designing non-grid based models lies in (1) handling a variable number of neighbours and (2) aggregating the state information of multiple neighbours to obtain the interaction vector $p_t^{i}$. As illustrated in Fig~\ref{fig:mix_grid_pool}, the design choices of these modules can be categorized based on four factors: (a) neighbour input state, (b) input state embedding, (c) neighbour information aggregation strategy, and (d) aggregated vector embedding.


\textbf{Neighbour input state:} Non-grid based methods do not contain an implicit notion of the spatial position of neighbours with respect to the primary pedestrian, unlike the grid-based counterparts. Hence, almost all the existing designs in literature take as input the relative spatial position of the neighbours. Another popular input choice is the hidden-state of the neighbouring pedestrian \cite{Kosaraju2019SocialBiGATMT, Gupta2018SocialGS} as the hidden-state has the ability to encode information regarding the motion history of the corresponding pedestrian. Amirian \textit{et al.}\cite{Amirian2019SocialWL} models the neighbour states using interaction-centric geometric features like bearing angle between agents and distance of closest approach \cite{Kooij2014ContextBasedPP}. Ivanovic \textit{et al.} \cite{Ivanovic2018TheTP} takes as input the velocity of neighbours. In this work, we argue that inputting \textit{relative} velocity of neighbours is an important factor for reducing collisions in model predictions.

\textbf{Input state embedding:} The input states of the neighbours are usually embedded using an MLP. However, recent works \cite{Vemula2017SocialAM, Haddad2019SituationAwarePT} based on graph neural network \cite{Bruna2014SpectralNA} designs, embed the relative input states using an LSTM. Each connection of a primary pedestrian to his neighbour is modelled using a different LSTM. The LSTM helps to capture the evolution of the relative neighbour states, unlike the first-order MLP.

\textbf{Aggregation strategy:} One of the most important challenges of non-grid based models is to find the ideal strategy to aggregate the information of all the neighbours. Gupta \textit{et al.} \cite{Gupta2018SocialGS} proposed to aggregate the interaction information by applying a symmetric max-pool function on the LSTM hidden-states of the neighbouring pedestrians. Ivanovic \textit{et al.} \cite{Ivanovic2018TheTP} and Hasan {et al.} \cite{Hasan2018MXLSTMMT} utilized the symmetric sum-pooling function. 

A large body of works utilize the attention mechanism \cite{Vaswani2017AttentionIA, Bahdanau2014NeuralMT} to determine the weights of different neighbours in predicting the future trajectory. These weights can be either hand-crafted \cite{Fernando2018SoftH} or learnt in a data-driven manner \cite{Kosaraju2019SocialBiGATMT, Sadeghian2018SoPhieAA, Amirian2019SocialWL}. The attention mechanism can be applied multiple times to model higher-order spatial interactions. The commonly used data-driven attention mechanism is the design proposed for the Transformer architecture \cite{Vaswani2017AttentionIA}.

A simple baseline for aggregating neighbour information is to concatenate the neighbour embeddings. To tackle the issue of handling variable number of neighbours, we investigate the performance of the concatenation scheme by selecting the top-$k$ neighbours based on a defined criterion, (\textit{e.g.}, euclidean distance). Despite the simplicity, we demonstrate that the concatenation strategy performs at par with its sophisticated counterparts.

\textbf{Aggregated vector embedding:} The aggregated neighbour vector is usually passed through an MLP, with the exception of Ivanovic \textit{et al.} \cite{Ivanovic2018TheTP} passes the sum-pooled neighbour information through an LSTM, to obtain the interaction vector $p_t^i$. We argue that encoding the aggregated vector using LSTMs offers the advantage of modelling higher-order interactions in the temporal domain. In other words, the interaction module learns how the interaction representations evolve over time.

\begin{table*}[h]
    \centering
\begin{center}
\resizebox{0.96\textwidth}{!}{\begin{tabular}{ |c|c|c|c|c|c| } 
 \hline
 \textbf{Acronym (P-Q-R-S)} & Input (P) & Embed-I (Q) & Aggreg. (R) & Embed-II (S) & References \\ 
 \hline
 O-Grid  & Position & None & Grid & MLP & \textbf{O-LSTM} \cite{Alahi2016SocialLH}, \cite{Bartoli2018ContextAwareTP}, \cite{Pfeiffer2017ADM}\\
 
 S-Grid  & H-State & None & Grid & MLP & \textbf{S-LSTM} \cite{Alahi2016SocialLH}, \cite{Bartoli2018ContextAwareTP}, \cite{Zhao2019MultiAgentTF}, \cite{Bisagno2018GroupLG},  \cite{Varshneya2017HumanTP}, \cite{Lisotto2019SocialAS}, \cite{Lee2017DESIREDF}, \cite{Hasan2018MXLSTMMT} \\
 
  D-Grid  & Velocity & None & Grid & MLP & \textbf{Directional Pooling} [\textbf{Ours}] \\
 \hline
  D-MLP-Attn-MLP & Velocity & MLP & Attn & MLP & \cite{Shi2019PedestrianTP} \\
  
  S-MLP-Attn-MLP & H-State & MLP & Attn & MLP & \textbf{S-BiGAT} \cite{Kosaraju2019SocialBiGATMT}, \cite{Amirian2019SocialWL}, \cite{Fernando2018SoftH}, \cite{Sadeghian2018SoPhieAA}, \cite{Zhang2019SRLSTMSR}, \cite{Xu2018EncodingCI}, \cite{Li2020SocialWaGDATIT}, \cite{Huang2019STGATMS}, \cite{Yu2020SpatioTemporalGT} \\
  
  S-MLP-MaxP-MLP & H-State & MLP & MaxPool & MLP & \textbf{S-GAN} \cite{Gupta2018SocialGS} \\
  
  D-MLP-ConC-MLP &  Velocity & MLP & Concat & MLP & \cite{Tordeux2019PredictionOP}, \cite{Ma2016AnAI} \\
  
  D-MLP-SumP-LSTM & Velocity & MLP & SumPool & MLP & \textbf{Trajectron} \cite{Ivanovic2018TheTP} \footnote{Absolute velocity and positions are taken as input} \\
  
  O-LSTM-Att-MLP & Position & LSTM & Attn & MLP & \textbf{S-Attn} \cite{Vemula2017SocialAM}, \cite{Haddad2019SituationAwarePT} \\
  D-MLP-ConC-LSTM & Velocity & MLP & Concat & LSTM & \textbf{DirectConcat} [\textbf{Ours}] \\
 \hline
\end{tabular}}
\end{center}
    \caption{Model Acronyms: Acronyms for the various designs of interaction modules. We observe that most of the existing interaction encoder designs fall under our defined categorization.}
    \label{tab:acroandref}
\end{table*}

For brevity, the interaction modules are denoted using acronyms based on their designs. The acronyms are of the form \textbf{P-Q-R-S} where \textbf{P} denotes the input to the module, \textbf{Q} denotes the state embedding module, \textbf{R} denotes the information aggregation mechanism and \textbf{S} denotes aggregated vector embedding module. The choices for each of these components is provided in Table~\ref{tab:acroandref}. The table also illustrates how our categorization encompasses the popular designs on NN-based interaction modules in literature. 

\textbf{DirectConcat} \\
Equivalent to our proposed \texttt{D-Grid}, we now describe its non-grid counterpart \textit{DirectConcat}. Grid-based models, based on their design, implicitly consider only those neighbours that are within the grid constructed around the primary pedestrian. We argue that modelling interactions of all pedestrians (even those far away) can lead to the model learning spurious correlations. Thus, we propose to consider only the top-$k$ neighbours closest to the primary pedestrian. We will demonstrate in the experimental section that if $k$ is set to a large value, \textit{i.e.} if the model considers all pedestrians in the scene, the model deteriorates in its ability to learn collision avoidance.

Similar to aggregating the obtained directional grid by flattening the obtained grid, in \textit{DirectConcat} we propose to concatenate the relative-velocity and relative-position embeddings of top-$k$ neighbours. This preserves the unique identity of the neighbours as compared to mixing the different embeddings like in max-pooling \cite{Gupta2018SocialGS} or sum-pooling \cite{Ivanovic2018TheTP}. Finally, we pass the aggregated vector through an LSTM as compared to an MLP. This design choice helps to model higher-order spatio-temporal interactions better and is more robust to noise in the real-world measurements as LSTM controls the evolution of the interaction vector. We demonstrate in the experimental section that indeed the LSTM embedding further helps to improve the collision metric. By design, \textit{DirectConcat} falls under the \texttt{D-MLP-ConC-LSTM} architecture of our categorization.  We will use the terms \textit{DirectConcat} and \texttt{D-MLP-ConC-LSTM} interchangeably.


\subsection{Forecasting Model}

We now describe the rest of the components of the forecasting model. To claim that a particular design of the interaction module is superior, it is essential to keep the rest of the forecasting model components constant. Only then we can be sure that it was the interaction module design that boosted performance, and not one of the extra added components. We choose the time-sequence encoder to be an LSTM due to its capability to handle varying input length and capture long-term
dependencies. Moreover, most works have LSTMs as their base motion-encoding architecture. 

The rest of the architecture we describe now is identical for all the methods described in the previous section. The state of person $i$ at time-step $t$, $\textbf{s}^t_i$, is embedded using a single layer MLP to get the state embedding $e^t_i$. We represent each person's state using his/her velocity, as switching the input representation from absolute coordinates to velocities increases the generalization power of sequence encoder. We obtain the interaction vector $p^t_i$ of person $i$ from the interaction encoder. We concatenate the interaction vector with the velocity embedding and provide the resultant vector as input to the sequence-encoding module. Mathematically, we obtain the following recurrence: 
\begin{align} \label{eq:LSTM_main}
    e^t_{i} &= \phi(\textbf{v}^t_i; W_{emb}), \\
    h^t_{i} &= LSTM(h^{t-1}_i, [e^t_{i}; p^{t}_{i}]; W_{encoder}),
\end{align}
where $\phi$ is the embedding function, $W_{emb}, W_{encoder}$ are the weights to be learned. The weights are shared between all persons in the scene.

The hidden-state of the LSTM at time-step $t$ of pedestrian $i$ is then used to predict the distribution of the velocity at time-step $t + 1$. Similar to Graves \cite{Graves2013GeneratingSW}, we output a bivariate Gaussian distribution parametrized by the mean  $\mu_i^{t+1} = (\mu_x,\mu_y)_i^{t+1}$, standard deviation $\sigma_i^{t+1} = (\sigma_x,\sigma_y)_i^{t+1}$ and correlation coefficient $\rho_i^{t+1}$:

\begin{equation}
    [\mu_i^{t}, \sigma_i^{t}, \rho_i^{t}] = \phi_{dec}(h_i^{t-1}, W_{dec}),
\end{equation}
where $\phi_{dec}$ is modelled using an MLP and $W_{dec}$ is learned. 

\textit{Training:} All the parameters of the forecasting model are learned by minimizing the negative log-likelihood (NLL) loss:
\begin{equation}
   \mathcal{L}_i(w) = - \sum_{t = T_{obs} + 1}^{T_{pred}} \log(\mathbb{P}(\textbf{v}_i^t | \mu_i^{t}, \sigma_i^{t}, \rho_i^{t})).
\end{equation}

Contrary to the general practice of training the model by minimizing the NLL loss for all the trajectories in the training dataset, we minimize the loss for only the primary pedestrian (defined in the next section) in each scene of the training dataset. We will demonstrate how this training procedure helps the model to better capture social interactions in the experimental section.

\textit{Testing:} During test time, till time-step $T_{obs}$, we provide the ground truth position of all the pedestrians as input to the forecasting model. From time $T_{obs + 1}$ to $T_{pred}$, we use the predicted position (derived from the predicted velocity) of each pedestrian as input to the forecasting model and predict the future trajectories of all the pedestrians.

\begin{figure}[h]
\centering
\begin{subfigure}[h]{0.48\textwidth}
\includegraphics[width=0.99\textwidth]{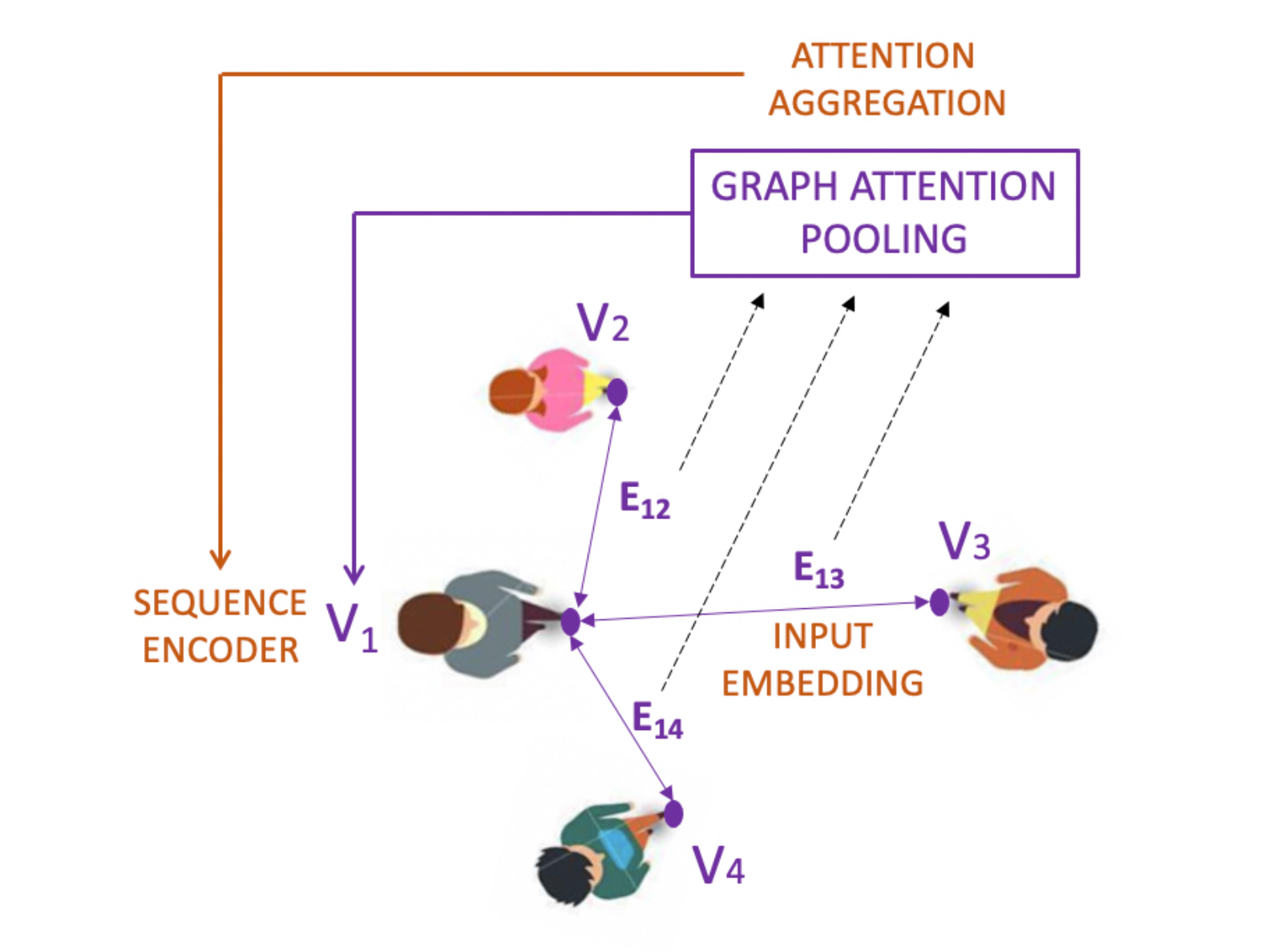}
\end{subfigure}
\centering 
\caption{Illustration of Graph neural networks (purple) as a special case of our data-driven pipeline (brown). Each vertex $V_i$ is modelled using a sequence encoder, the neighbour edges $E_{ij}$ correspond to our input embeddings which are aggregated via attention mechanism. The resulting interaction vector is provided as input to the sequence encoder (vertex $V_{i}$).}
\label{fig:GNN}
\end{figure}

\subsubsection{Equivalence to Graph Neural Networks}

Recently, graph neural networks (GNNs) have become popular for forecasting human motion. In the GNN setup,  each pedestrian is represented as a node/vertex $V_{i}$ and two interacting pedestrians are connected via an edge $E_{ij}$. $V_{i}$ models the sequence representation of the associated pedestrian and edge $E_{ij}$ updates according to the interactions between the associated pedestrians. We show an equivalence between dynamic-interaction-based graph neural networks and our proposed LSTM-based pipeline with \texttt{S-X-Attn-MLP} (where \texttt{X} $\in$ \{\texttt{MLP}, \texttt{LSTM}\}) interaction encoding scheme, visually illustrated in Figure~\ref{fig:GNN}. Without loss of generality, let pedestrian $i$ be the primary pedestrian. Vertex $V_{i}$ is modelled using an LSTM sequence encoder. Edge $E_{ij}$ takes as input the state of the neighbours and updates over time using an MLP or LSTM (\textit{input state embedding}). At each time-step, the information of all connected edges is aggregated using attention mechanism (\textit{aggregation strategy}), popularly referred to as GAT-pooling \cite{Velickovic2018GraphAN} in GNN literature. Finally the aggregated vector is optionally passed through an MLP to obtain the interaction vector $p_{i}$ which is the input to the LSTM sequence encoder for $V_{i}$. Social-BiGAT \cite{Kosaraju2019SocialBiGATMT} utilizes the \texttt{S-MLP-Attn-MLP} design, Social Attention \cite{Vemula2017SocialAM} utilizes the \texttt{O-LSTM-Attn-MLP} design while recently, STAR \cite{Yu2020SpatioTemporalGT} utilizes the \texttt{S-MLP-Attn-MLP} design with the sequence encoder for vertex $V_i$ being a Transformer \cite{Vaswani2017AttentionIA}.

\subsection{Explaining trajectory forecasting models}
Trajectory forecasting models are deployed in many safety-critical applications like autonomous systems. In such scenarios, it becomes really important to gain insight into the decision-making of so-called `blackbox' neural networks. Several works in literature attempt to explain the rationale behind the NN decisions \cite{Simonyan2015VeryDC, Bach2015OnPE, Springenberg2015StrivingFS, Sundararajan2017AxiomaticAF, JMLR:v20:18-540}. Out of these techniques, Layer-wise Relevance Propagation (LRP) is one of the most prominent methods in explainable machine learning.

LRP re-distributes the model output score to each of the input variables indicating the extent to which they contribute to the output. LRP works by reverse-propagating the prediction through the network by means of heuristic propagation rules that apply to each layer of a neural network \cite{Bach2015OnPE}. These propagation rules are based on a local conservation principle: the net quantity or relevance, received by any higher layer neuron is redistributed in the same amount to neurons of the layer below. Mathematically, if $j$ and $k$ are indices for neurons in two consecutive layers, and denoting by $R_{j \shortrightarrow k}$ the relevance flowing between two neurons, we have the equations:
\begin{align}
    \Sigma_{j} R_{j \shortrightarrow k} & = R_{k} \\
    R_{j} & = \Sigma_{k} R_{j \shortrightarrow k}
\end{align}
On applying the local conservation principle across all the layers, we obtain global conservation of the output score when reverse propagated back to the inputs. Recently, Arras \textit{et al.} \cite{Arras2019ExplainingAI} have shown that the principle of LRP can also be applied to LSTMs.

LRP has largely been explored in the domain of model classification \textit{i.e.} the outputs are classification scores. In this work, we utilize LRP to determine on which neighbours (via the input pooling map) and past velocities (via the input velocity embedding) of the primary pedestrian our model focuses on, when regressing to the next predicted velocity. We achieve this by reverse-propagating both the x-component $v_x$ as well as y-component $v_y$ of predicted velocity ($\mathbf{v_{pred}} = (v_x, v_y)$) and adding the obtained input relevance scores. To the best of our knowledge, we are the first work to empirically demonstrate that LRP provides reasonable explanations when extended to the regression task of trajectory forecasting. Moreover, the LRP technique is generic and can be applied on top of any trajectory forecasting network to analyze its predictions.

\begin{figure*}[t]
\begin{subfigure}[t]{0.24\textwidth}
\includegraphics[width=0.95\textwidth]{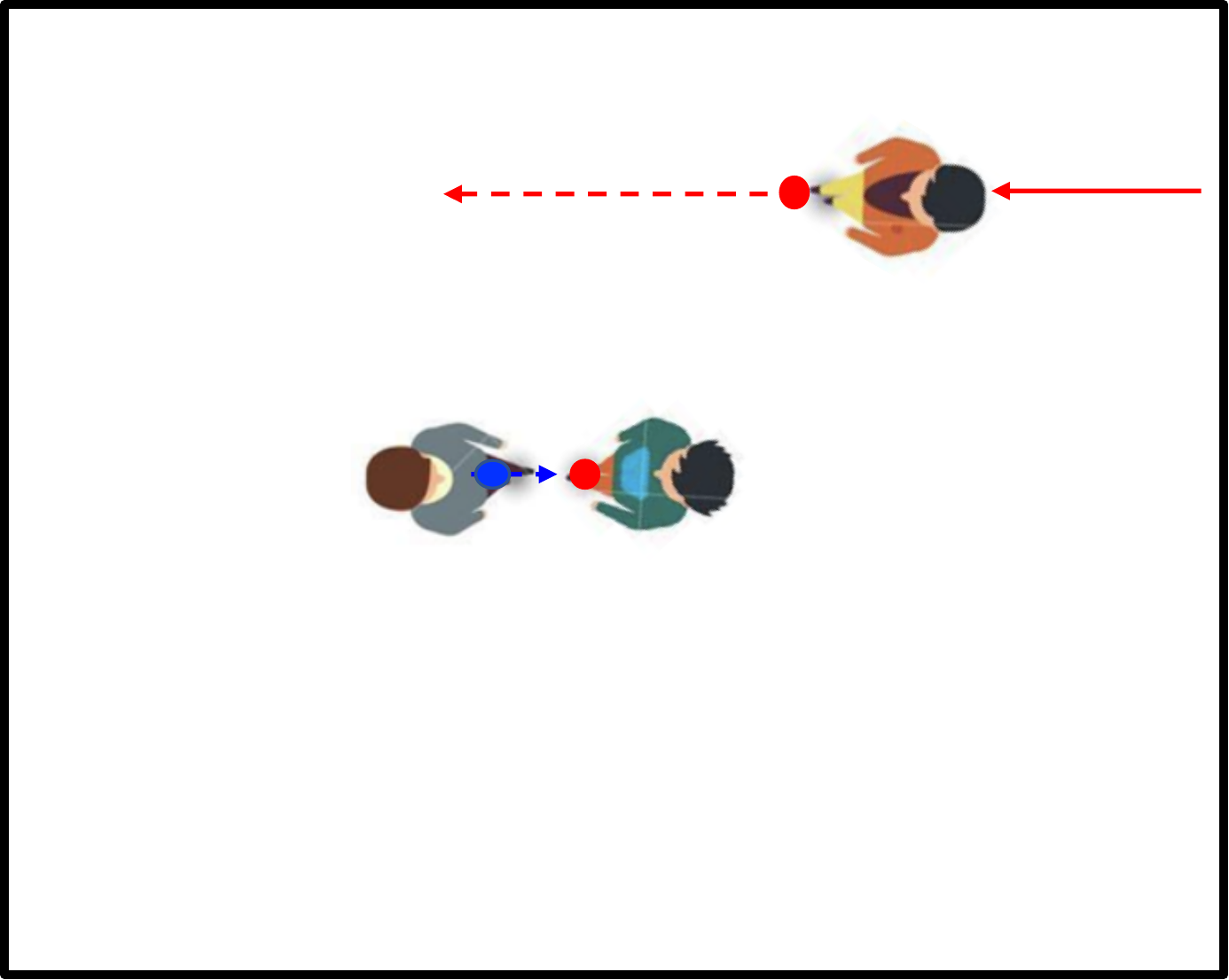}
\caption{Static}
\end{subfigure}
\hfill
\begin{subfigure}[t]{0.24\textwidth}
\includegraphics[width=0.95\textwidth]{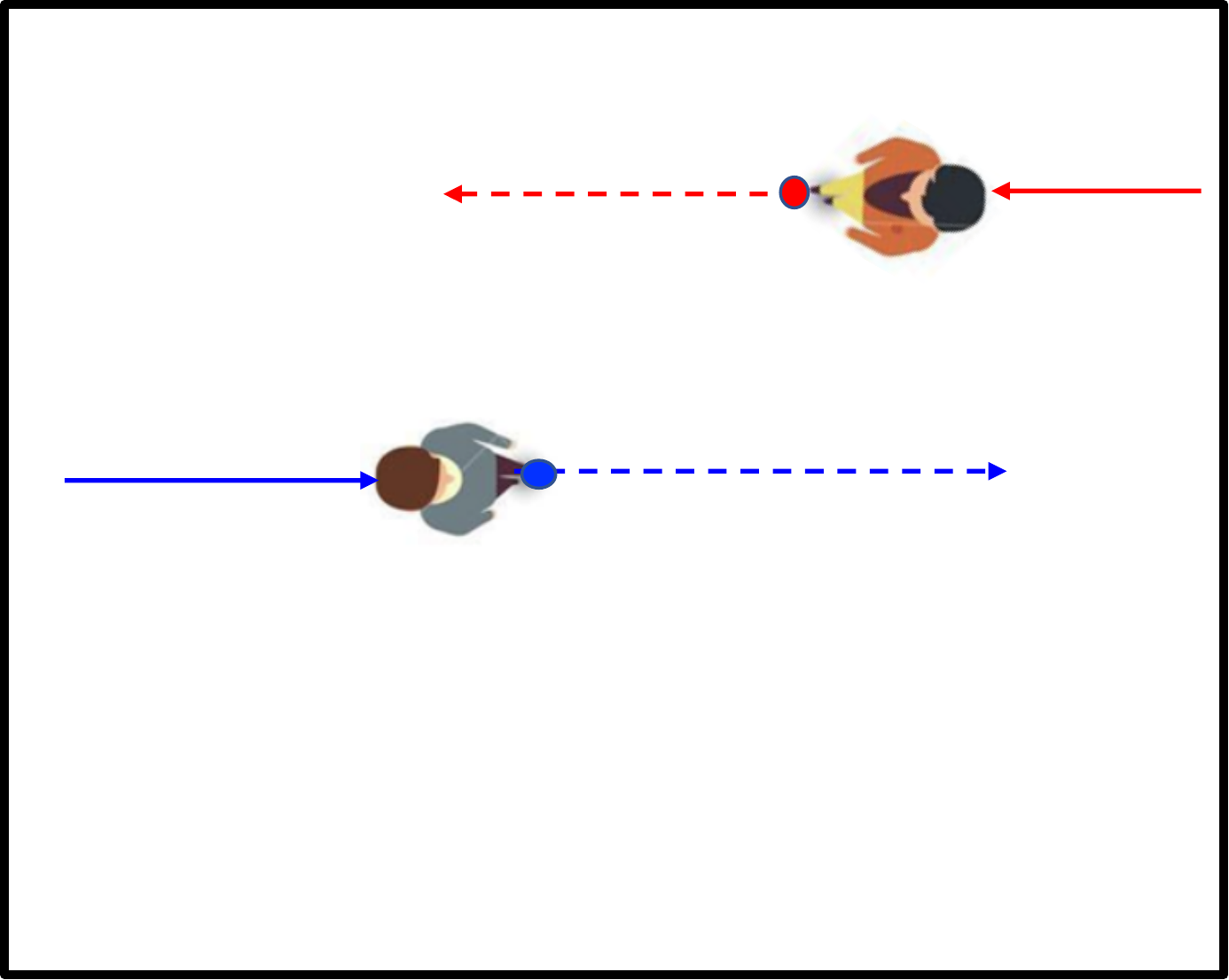}
\caption{Linear}
\end{subfigure}
\begin{subfigure}[t]{0.24\textwidth}
\includegraphics[width=0.95\textwidth]{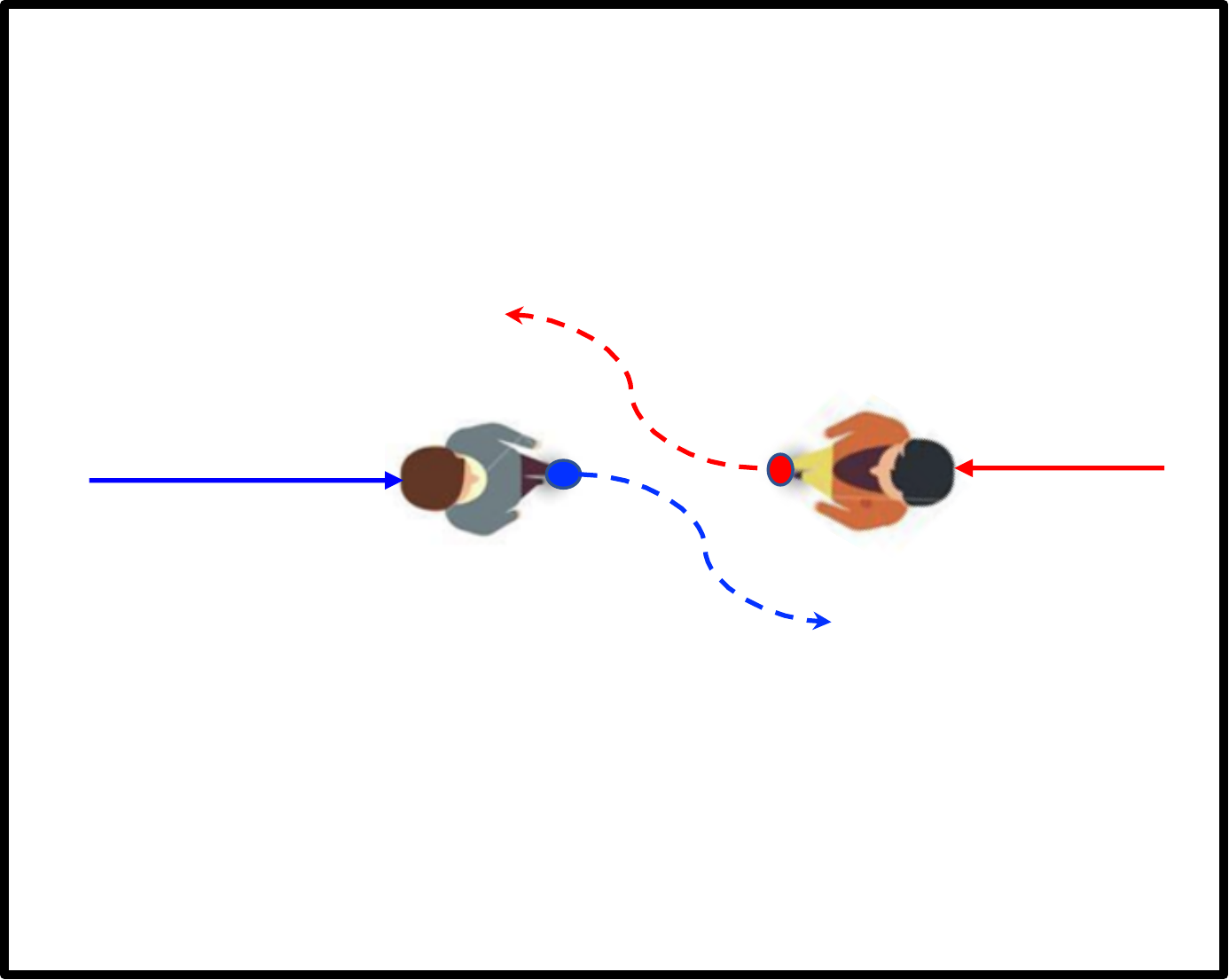}
\caption{Interacting}
\end{subfigure}
\hfill
\begin{subfigure}[t]{0.24\textwidth}
\includegraphics[width=0.95\textwidth]{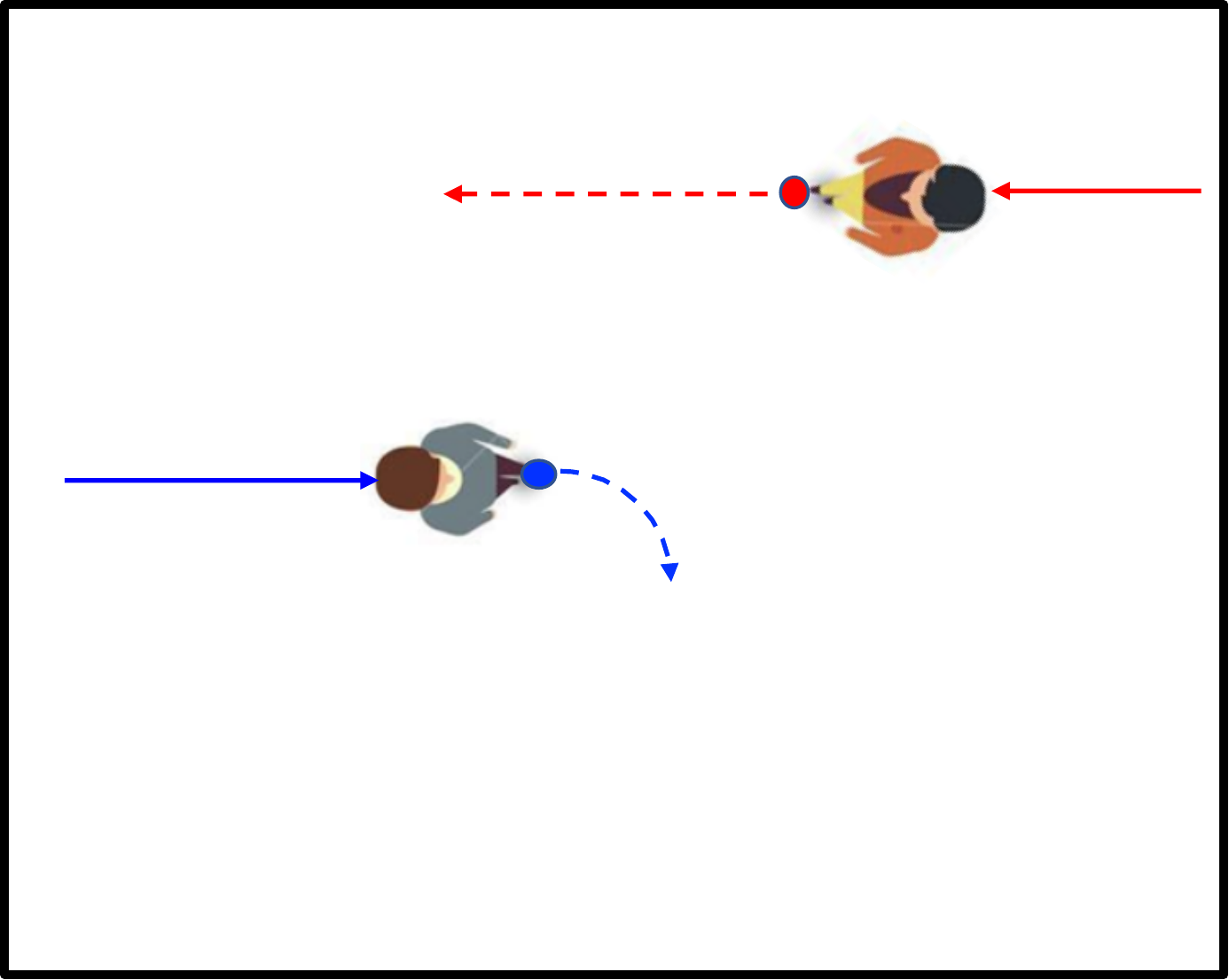}
\caption{Non-Interacting}
\end{subfigure}
\caption{Visualization of our high-level defined trajectory categories}
\label{fig:overall_cat}
\end{figure*}

\begin{figure*}[t]
\begin{subfigure}[h]{0.24\textwidth}
\includegraphics[width=0.95\textwidth]{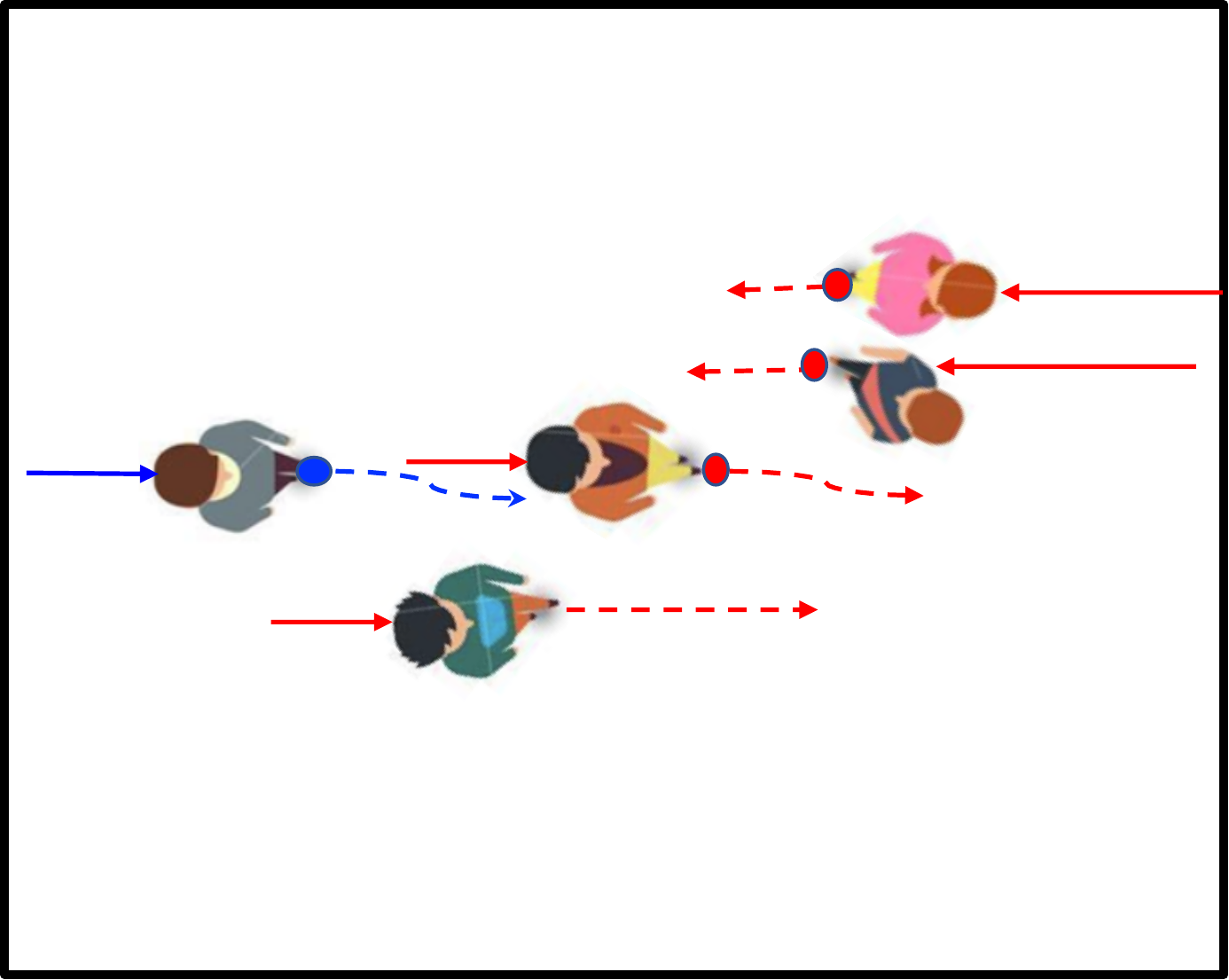}
\caption{Leader Follower}
\end{subfigure}
\hfill
\begin{subfigure}[h]{0.24\textwidth}
\includegraphics[width=0.95\textwidth]{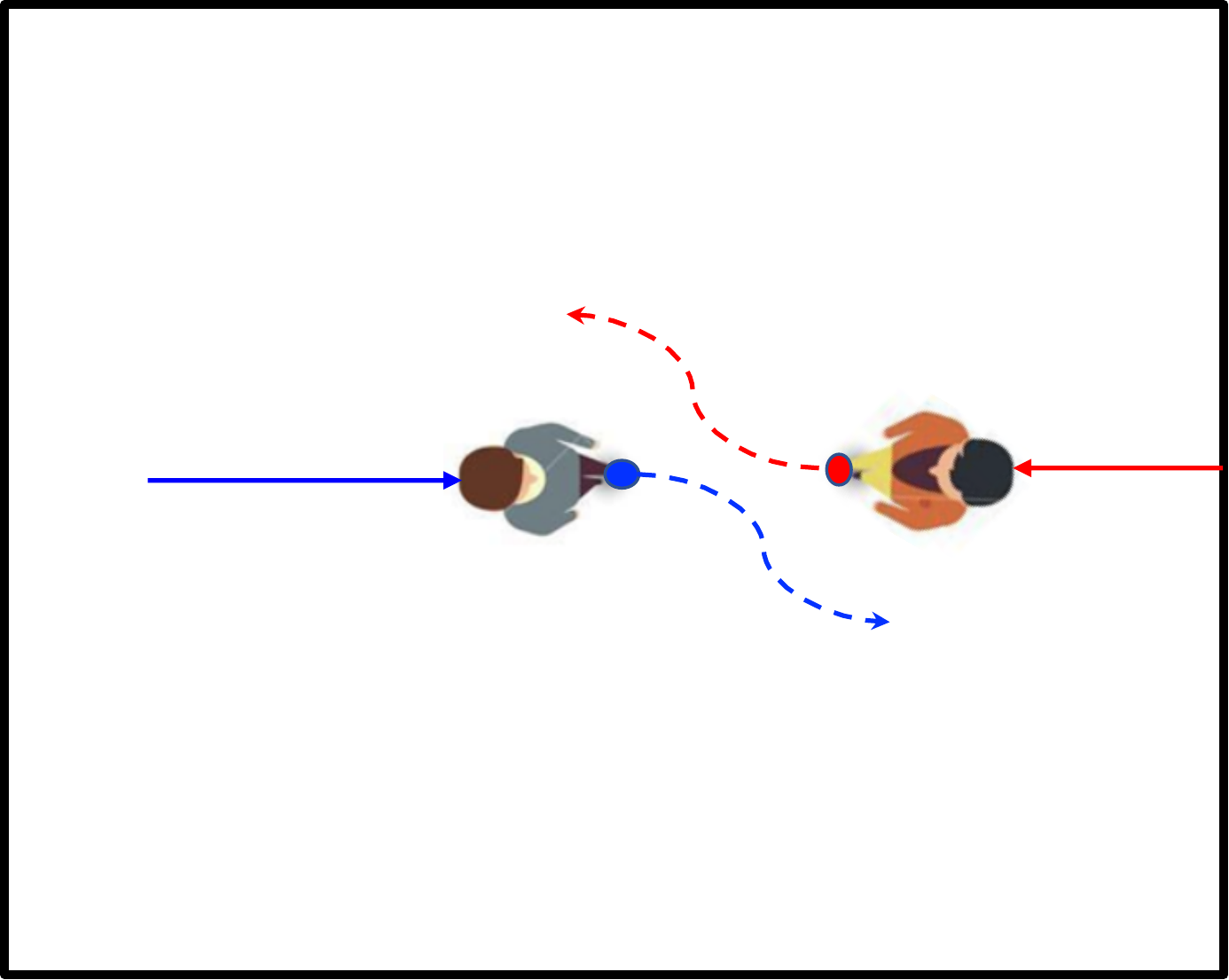}
\caption{Collision avoidance}
\end{subfigure}
\begin{subfigure}[h]{0.24\textwidth}
\includegraphics[width=0.95\textwidth]{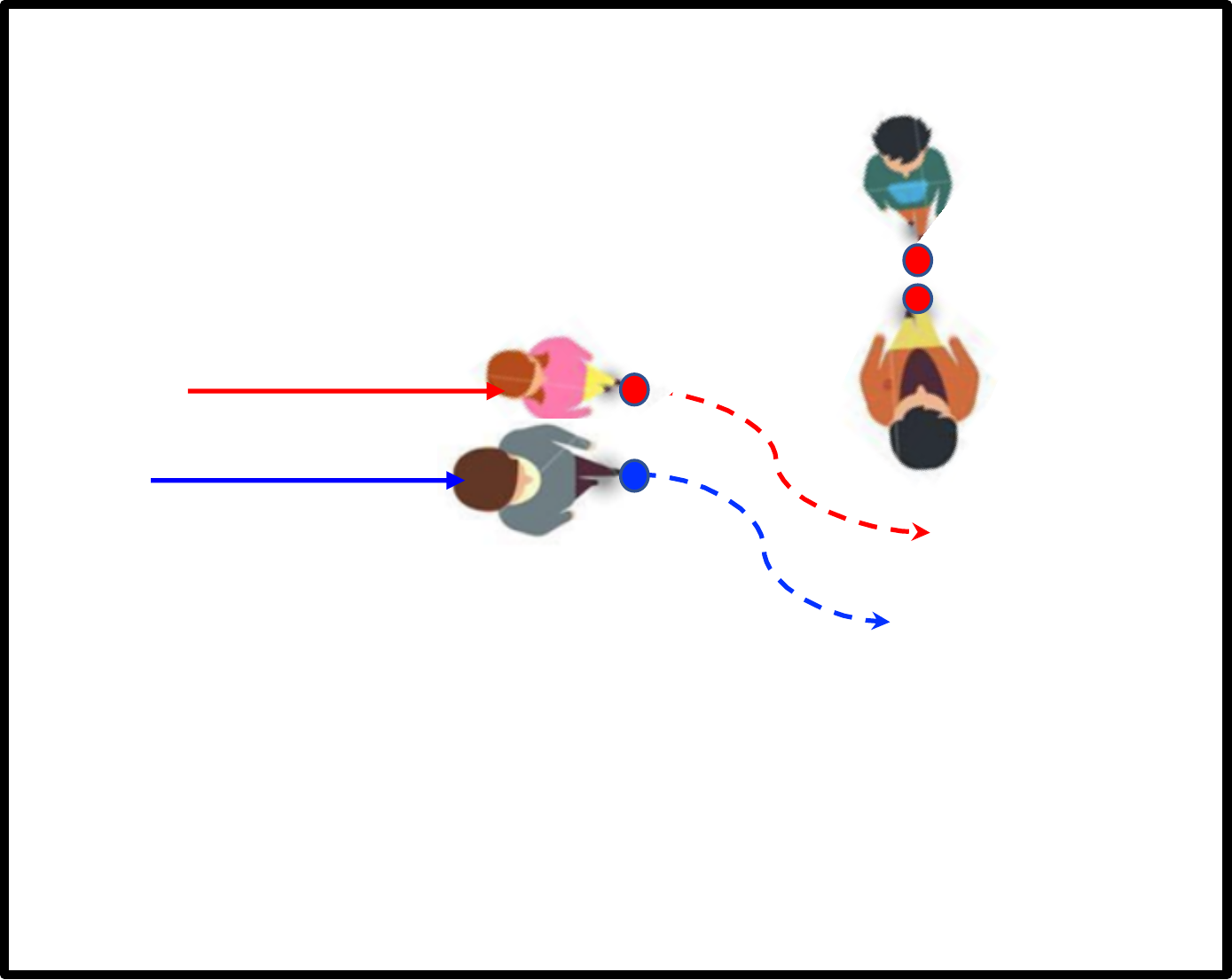}
\caption{Group}
\end{subfigure}
\hfill
\begin{subfigure}[h]{0.24\textwidth}
\includegraphics[width=0.95\textwidth]{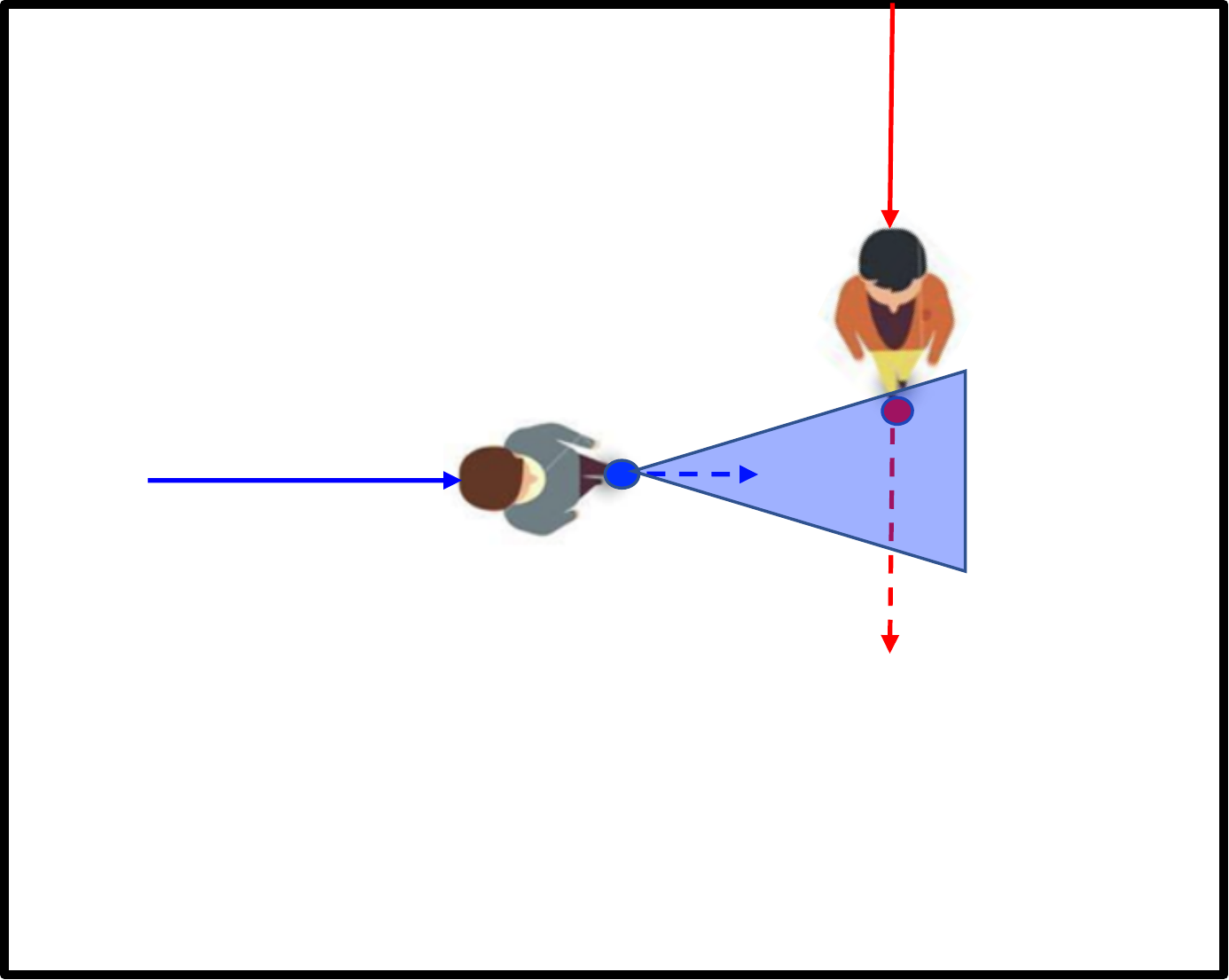}
\caption{Others}
\end{subfigure}

\caption{Visualization of our Type III interactions commonly occurring in real world crowds.}
\label{fig:type_iii}
\end{figure*}

\begin{figure}[h]
\centering
\includegraphics[width=0.48\textwidth]{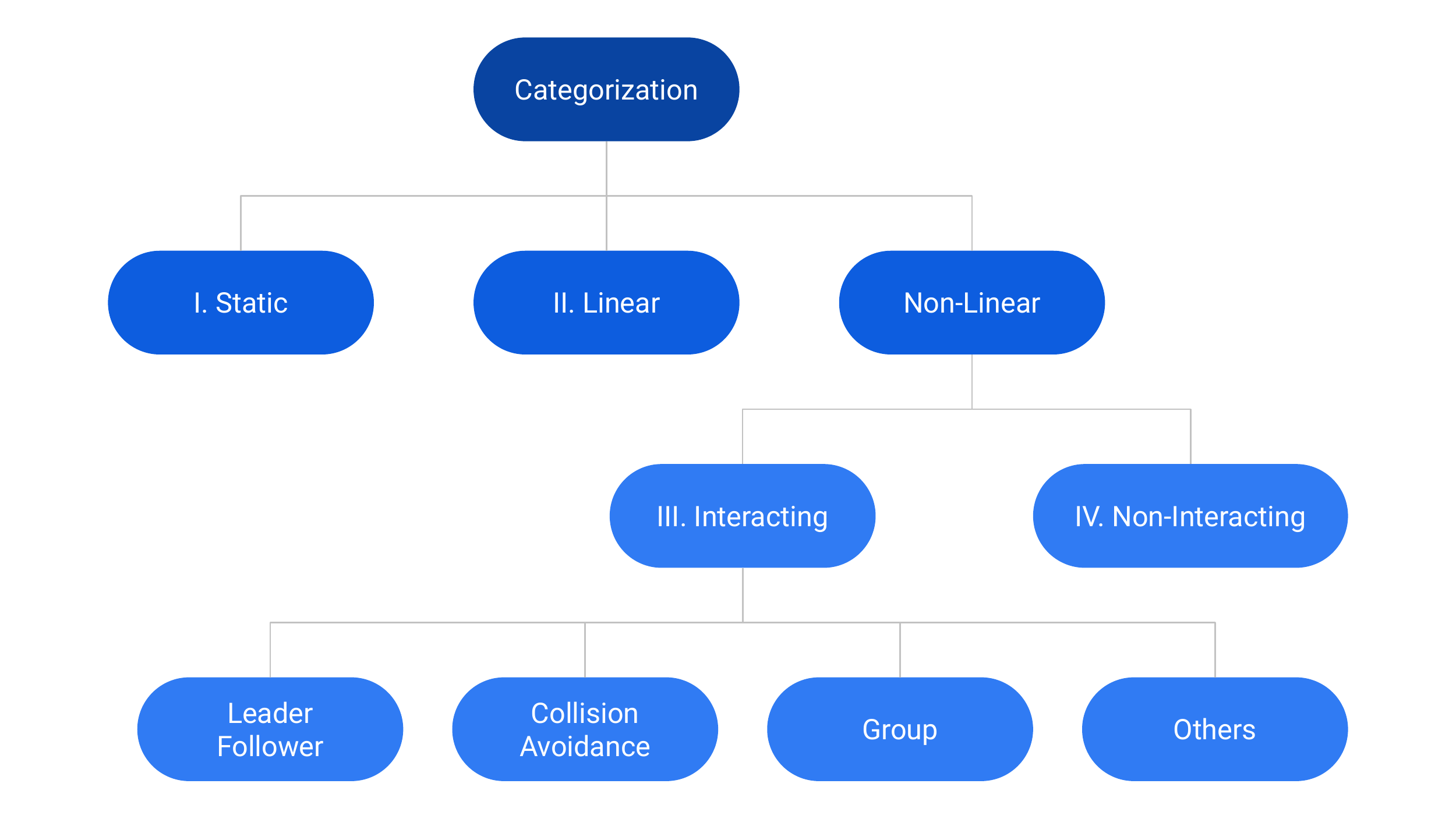}
\caption{Our proposed hierarchy for trajectory categorization. Using our defined trajectory categorization, we construct the \textit{TrajNet++} benchmark by sampling trajectories corresponding largely to `Type III: Interacting' category.}
\label{fig:traj_categ}
\end{figure}

\section{TrajNet++: A Trajectory Forecasting Benchmark}

In this section, we present \textit{TrajNet++}, our interaction-centric human trajectory forecasting benchmark. To demonstrate the efficacy of a trajectory forecasting model, the standard practice is to evaluate these models against baselines on a standard benchmark. However, current methods have been evaluated on different subsets of available data without proper sampling of scenes in which social interactions occur. In other words, a data-driven method cannot learn to model agent-agent interactions if the benchmark comprises primarily of scenes where the agents are static or move linearly. Therefore, our benchmark comprises largely of scenes where social interactions occur. To this extent, we propose the following trajectory categorization hierarchy. 


\begin{figure*}[h]
\centering
\begin{subfigure}[h]{0.24\textwidth}
\includegraphics[width=0.95\textwidth]{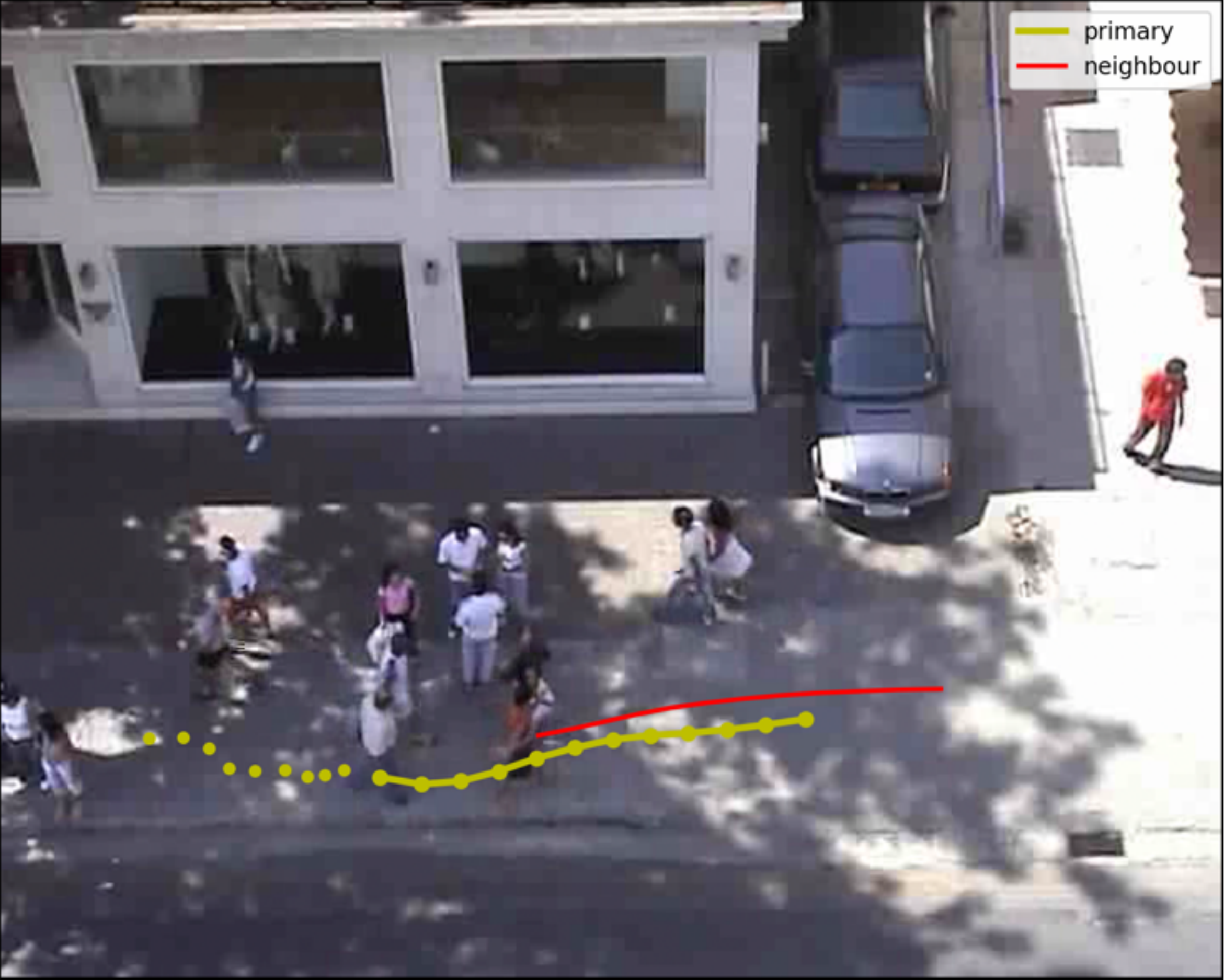}
\caption{Leader Follower}
\end{subfigure}
\begin{subfigure}[h]{0.24\textwidth}
\includegraphics[width=0.95\textwidth]{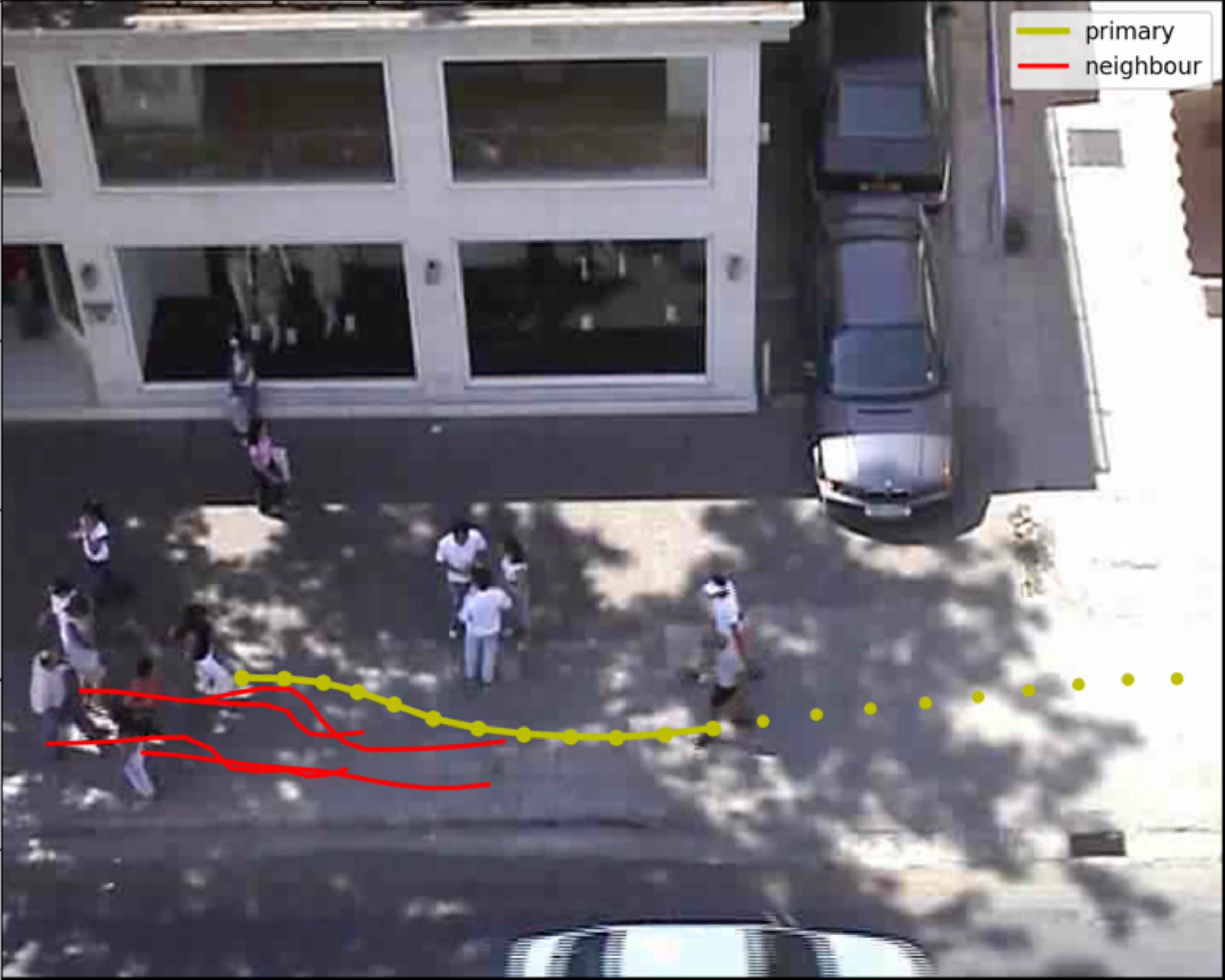}
\caption{Collision Avoidance}
\end{subfigure}
\vspace{0.3 em}
\begin{subfigure}[h]{0.24\textwidth}
\includegraphics[width=0.95\textwidth]{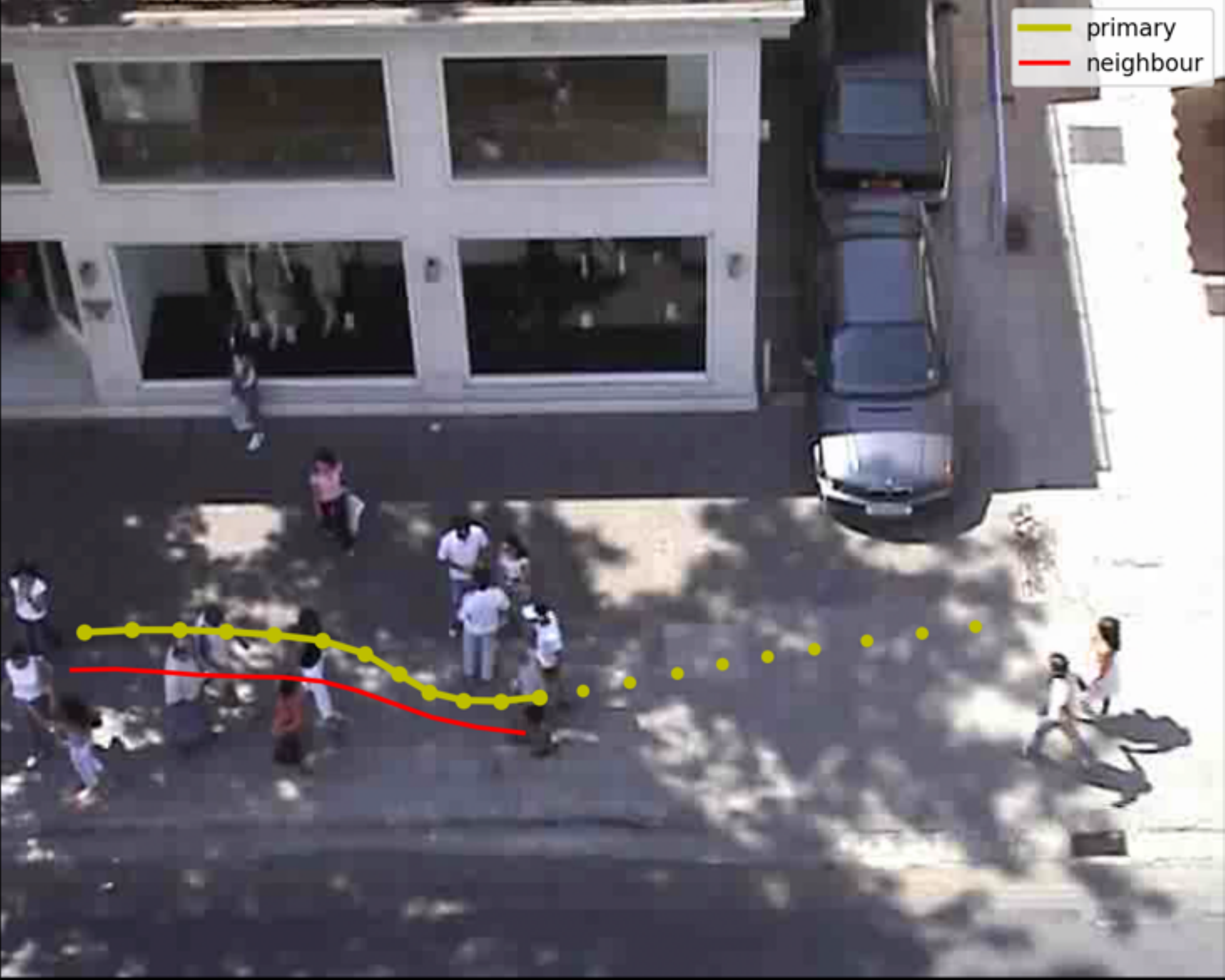}
\caption{Group}
\end{subfigure}
\begin{subfigure}[h]{0.24\textwidth}
\includegraphics[width=0.95\textwidth]{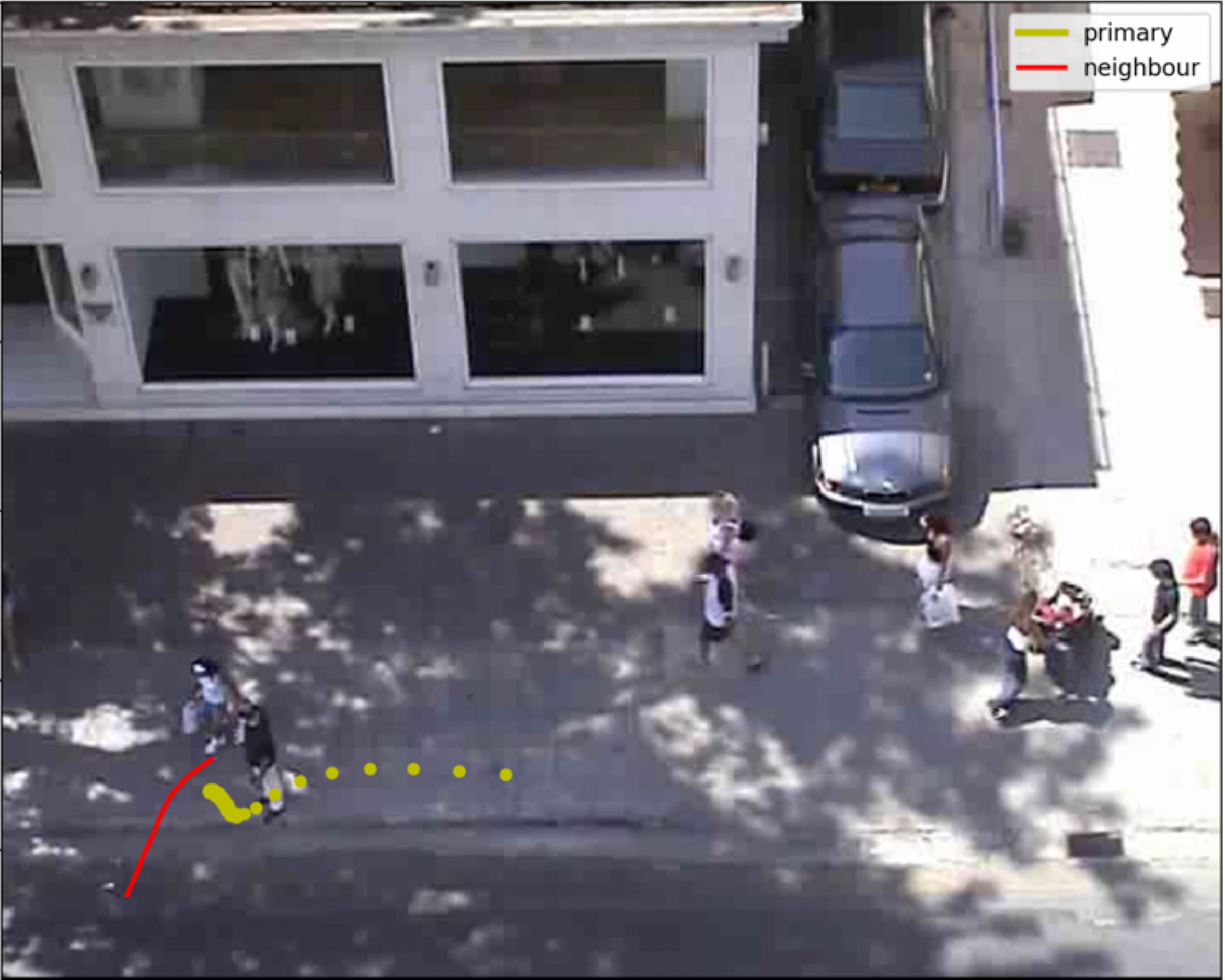}
\caption{Others}
\end{subfigure}
\centering
\caption{Sample scenes from our benchmark. In each of the samples, we illustrate a different social interaction between the primary pedestrian (yellow) and the corresponding interacting neighbours (red) in real world datasets.}
\label{fig:realworld_type3}
\end{figure*}

\subsection{Trajectory Categorization}
We provide a detailed trajectory categorization (Fig~\ref{fig:traj_categ}). This detailed categorization helps us not only to better sample trajectories for TrajNet++ dataset but also glean insights into the model performance in diverse scenarios, \textit{i.e.}, to verify whether the model captures all the different kinds of interactions.

To aid our categorization, we introduce the notion of a \textit{primary pedestrian} as a reference pedestrian with respect to which we categorize scenes. Each scene has a primary pedestrian whose motion we want to forecast. We refer to the other pedestrians in the scene as \textit{neighbouring pedestrians.}

We explain in detail our proposed hierarchy for trajectory categorization (Fig~\ref{fig:traj_categ}). We also provide example scenarios for the same in  Fig~\ref{fig:overall_cat}:
\begin{enumerate}
\item \textbf{Static (Type I)}: If the euclidean displacement of the primary pedestrian in the scene is less than a specific threshold.
\item \textbf{Linear (Type II)}: If the trajectory of the primary pedestrian can be \textit{correctly forecasted} with the help of an Extended Kalman Filter (EKF). A trajectory is said to be \textit{correctly forecasted} by EKF if the FDE between the ground truth trajectory and forecasted trajectory is less than a specific threshold. 
\end{enumerate}
The rest of the scenes are classified as `Non-Linear'. We further divide non-linear scenes into Interacting (Type III) and Non-Interacting (Type IV).
\begin{enumerate}
  \setcounter{enumi}{2}
\item \textbf{Interacting (Type III)}: These correspond to scenes where the primary trajectory undergoes social interactions. For a detailed categorization coherent with commonly observed social interactions, we divide interacting trajectories into the following sub-categories (shown in Fig~\ref{fig:type_iii}).
\end{enumerate}
\begin{enumerate}[label=(\alph*)]
\item \textbf{Leader Follower [LF] (Type IIIa)}: Leader follower phenomenon refers to the tendency to follow pedestrians going in relatively the same direction. The follower tends to regulate his/her speed and direction according to the leader. If the primary pedestrian is a follower, we categorize the scene as Leader Follower.
\item \textbf{Collision Avoidance [CA] (Type IIIb)}: Collision avoidance phenomenon refers to the tendency to avoid pedestrians coming from the opposite direction. We categorize the scene as Collision avoidance if the primary pedestrian to be involved in collision avoidance.
\item \textbf{Group (Type IIIc)}: The primary pedestrian is said to be a part of a group if he/she maintains a close and roughly constant distance with at least one neighbour on his/her side during prediction.
\item \textbf{Other Interactions (Type IIId)}: Trajectories where the primary pedestrian undergoes social interactions other than LF, CA and Group. We define \textit{social interaction} as follows: We look at an angular region in front of the primary pedestrian. If any neighbouring pedestrian is present in the defined region at any time-instant during prediction, the scene is classified as having the presence of social interactions. 
\end{enumerate}
\begin{enumerate}
\setcounter{enumi}{3}
\item \textbf{Non-Interacting (Type IV)}: If a trajectory of the primary pedestrian is non-linear and undergoes no social interactions during prediction. 
\end{enumerate}

Using our defined trajectory categorization, we construct the \textit{TrajNet++} benchmark by sampling trajectories corresponding mainly to the Type III category. Moreover, having many Type-I scenes in a dataset can hamper the training of the model and result in misleading evaluation. Therefore, we remove such samples in the construction of our benchmark. The details of the categorization thresholds as well as the datasets which comprise our TrajNet++ benchmark are provided in the supplementary material. A few examples of our categorization in the real world are displayed in Fig~\ref{fig:realworld_type3}. In addition to comprising well-sampled trajectories, \textit{TrajNet++} provides an extensive evaluation system to understand model performance better.


\begin{figure}[h]
\centering
\begin{subfigure}[h]{0.48\textwidth}
\includegraphics[width=0.95\textwidth]{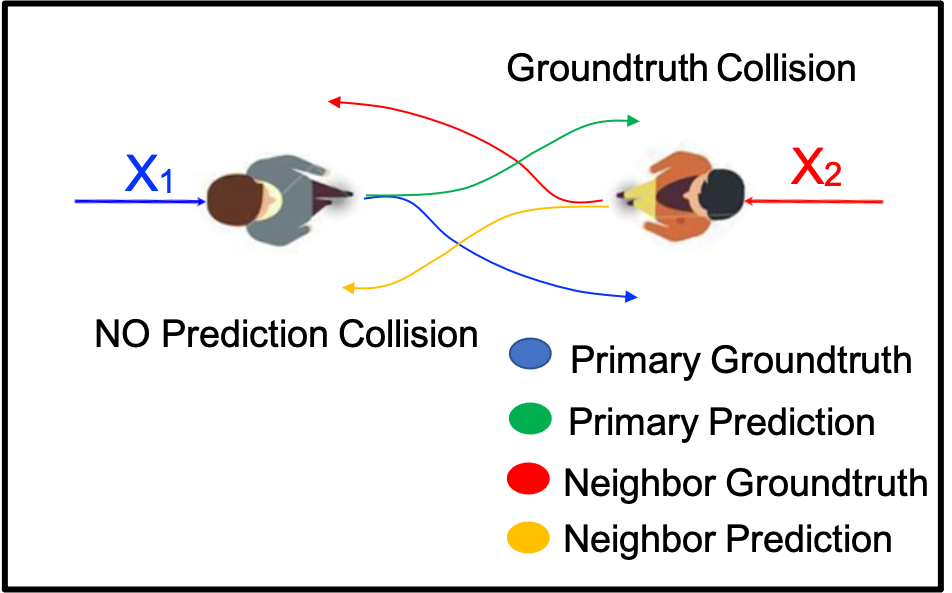}
\end{subfigure}
\caption{Visual illustration of our proposed collision metrics. In the example, the model prediction exhibits ground-truth collision (Col-II = 1) but no prediction collision (Col-I = 0).}
\label{fig:col_metrics}
\end{figure}

\subsection{Evaluation Metrics}\label{sec:eval_metrics}

\textbf{Unimodal Evaluation}: Unimodal evaluation refers to the evaluation of models that propose a single future mode for a given past observation. The most commonly used metrics of human trajectory forecasting in the unimodal setting are Average Displacement Error (ADE) and Final Displacement Error (FDE) defined as follows:
\begin{enumerate}
  \item \textbf{Average Displacement Error (ADE)}: Average $L_{2}$ distance between ground truth and model prediction overall predicted time steps. 
  \item \textbf{Final Displacement Error (FDE)}: The distance between the predicted final destination and the ground truth final destination at the end of the prediction period $T_{pred}$.
\end{enumerate}

These metrics essentially define different distance measures between the forecasted trajectory and the ground truth trajectory. With respect to our task, one of the most important aspects of human behavior in crowded spaces is collision avoidance. To ensure that models forecast feasible collision-free trajectories, we propose two new collision-based metrics in our framework (see Fig~\ref{fig:col_metrics}):
\begin{enumerate}
   \setcounter{enumi}{2}
   \item \textbf{Collision I - Prediction collision (Col-I)}: This metric calculates the percentage of collision between the primary pedestrian and the neighbors in the \textit{forecasted future} scene. This metric indicates whether the predicted model trajectories collide, \textit{i.e.}, whether the model learns the notion of collision avoidance.   
   \item \textbf{Collision II - Groundtruth collision (Col-II)}:  This metric calculates the percentage of collision between the primary pedestrian's prediction and the neighbors in the \textit{groundtruth future} scene.
\end{enumerate}
We want to stress further the importance of the collision metrics in the unimodal setup. As mentioned earlier, human motion is multimodal. A model may forecast a physically-feasible future, which is different from the actual ground truth. Such a physically-feasible prediction can result in a large ADE/FDE, which can be misleading. Our Col-I metric can help overcome this limitation of ADE/FDE metrics and provides a solution to measure `physical feasibility' of a prediction (aversion to a collision in this case). Col-II metric indicates whether the model understood the intention of the neighbours and predicted the desired trajectory mode indicated by fewer collisions with neighbours in ground truth. We believe our proposed collision metrics are an important step towards capturing the understanding of the model of human social etiquette in crowds. \\ 


\textbf{Multimodal Evaluation}: For models performing multimodal forecasting, \textit{i.e.}, outputting a future trajectory distribution, we provide the following metrics to measure their performance:
\begin{enumerate}
   \setcounter{enumi}{4}
   \item \textbf{Top-k ADE}: Given $k$ output predictions for an observed scene, this metric calculate the ADE of the prediction \textit{closest} to the groundtruth trajectory, similar in spirit to Variety Loss \cite{Gupta2018SocialGS}. 
   \item \textbf{Top-k FDE}: Given $k$ output predictions for an observed scene, this metric calculate the FDE of the prediction \textit{closest} to the groundtruth trajectory, similar in spirit to Variety Loss \cite{Gupta2018SocialGS}.
\end{enumerate}

For the Top-k metrics, we propose $k$ be small (3 as opposed to 20) as a model outputting uniformly-spaced predictions, irrespective of the input observation, can result in a much lower Top-20 ADE/FDE.

\begin{enumerate}
   \setcounter{enumi}{6}
   \item \textbf{Average NLL}: This metric was proposed by Boris \textit{et. al.} \cite{Ivanovic2018TheTP}. At each prediction step, the authors utilize a Kernel Density Estimate (KDE) \cite{Parzen1962ONEO}. From these estimates, the log-likelihood of ground truth trajectory is computed at each time step and is subsequently averaged over the prediction horizon. This metric provides a good indication of the probability of the ground truth trajectory in the model prediction distribution. 
\end{enumerate}

\section{Experiments}

In this section, we perform extensive experimentation on both TrajNet++ synthetic and real-world datasets to understand the efficacy of various interaction module designs for human trajectory forecasting. Moreover, we demonstrate how our proposed metrics help to provide a complete picture of model performance. 


\subsection{Implementation Details}

The velocity of each pedestrian is embedded into a 64-dimensional vector. The dimension of the interaction vector is 256. The dimension of the goal direction vector is 64. For grid-based interaction encoding, we construct a grid of size $16 \times 16$ with a resolution of $0.6$ meters. The dimension of the hidden state of both the encoder LSTM and decoder LSTM is 128. As mentioned earlier, each pedestrian has its own encoder and decoder. The batch size is fixed to 8. We train using ADAM optimizer \cite{Kingma2015AdamAM} with a learning rate of 1e-3. We perform interaction encoding at every time-step. For the concatenation based models, we consider top-4 nearest neighbours based on euclidean distance unless stated otherwise. For the attention aggregation strategy, we utilize the attention mechanism proposed in the Transformer architecture \cite{Vaswani2017AttentionIA}.


Data augmentation is another technique that can help increase accuracy, which can get wrongly attributed to the interaction encoder. We use rotation augmentation as the data augmentation technique to regularize all the models.

\subsection{Interaction Models: Synthetic Experiments}

We utilize synthetic datasets to validate the efficacy of various interaction modules in a controlled setup. For the synthetic dataset, since ORCA has access to the goals of each pedestrian, we embed the goal-direction and concatenate it to the velocity embedding (see Eq~\ref{eq:LSTM_main}).

\begin{table}[t]
\centering
\resizebox{0.48\textwidth}{!}{
\begin{tabular}{ |c|c|c|c| } 
 \hline
\textbf{Model (Acronym)} &  \textbf{ADE/FDE} & \textbf{Col-I} & \textbf{Col-II} \\ 
 \hline
 \hline
 \multicolumn{4}{|c|}{\textbf{Grid based methods}}\\
\hline
 Vanilla                                                        & 0.32/0.62 & 19.0 & 7.1\\
 \hline
 O-Grid   \cite{Alahi2016SocialLH}                      & 0.27/0.52 & 11.7 & 4.9\\ 
 \hline
 S-Grid \cite{Alahi2016SocialLH}                       & 0.24/0.50 & \textbf{2.2} & \textbf{4.6}\\ 
 \hline
 D-Grid [\textbf{Ours}]                & 0.24/0.49 & \textbf{2.2} & 4.8\\ 
 \hline
 \hline
  \multicolumn{4}{|c|}{\textbf{Non-Grid based methods}}\\
\hline
  S-MLP-MaxP-MLP \cite{Gupta2018SocialGS}              & 0.27/0.52 & 6.4 & \textbf{5.2}\\
 \hline
  S-MLP-Attn-MLP \cite{Kosaraju2019SocialBiGATMT}      & 0.26/0.52 & 3.7 & 5.4\\ 
 \hline
 D-MLP-SumP-LSTM \cite{Ivanovic2018TheTP}      & 0.29/0.57 & 13.8 & 6.6\\  
 \hline
  O-LSTM-Attn-MLP \cite{Vemula2017SocialAM}      & 0.24/0.48 & 0.8 & 5.2 \\ 
 \hline
 D-MLP-MaxP-MLP   &        0.28/0.55 & 14.3 & 6.1 \\
 \hline
 D-MLP-Attn-MLP            & 0.27/0.52 & 8.1 & \textbf{5.0}  \\
 \hline
 D-MLP-ConC-MLP      & 0.25/0.50 & 1.3 & 5.6\\ 
  \hline
 D-MLP-ConC-LSTM [\textbf{Ours}]             & 0.24/0.48 & \textbf{0.6} & 5.3\\ 
 \hline
\end{tabular} }
\caption{Unimodal Comparison of interaction encoder designs when forecasting 12 future time-steps, given the previous 9 time-steps, on TrajNet++ synthetic dataset. Errors reported are ADE / FDE in meters, Col I / Col II reported in \%. We emphasize that our goal is to reduce Col-I without compromising distance-based metrics.}
\label{Synth_456_comparison}
\end{table}

Table~\ref{Synth_456_comparison} quantifies the performance of the different designs of interaction modules published in the literature on TrajNet++ synthetic dataset. It is very interesting to note how our proposed Col-I metric provides a more complete picture of model performance. Observing only the distance-based metrics, one might wrongly conclude that the methods are similar in performance, however, they do not indicate the ability of the model to learn social etiquette (collision avoidance in this case). In safety-critical scenarios, it is more important for a model to prevent collisions in comparison to minimizing ADE/FDE. 

\subsubsection{\textbf{Grid-Based Models}} our proposed \texttt{D-Grid} outperforms \texttt{O-Grid}, especially in terms of Col-I, \textit{i.e.}, \texttt{D-Grid} learns better to avoid collisions. It is interesting to note that even though the motion encoder (LSTM) has the potential to infer the relative velocity of neighbours over time, there is a significant difference in performance when we explicitly provide relative velocity of the neighbours as input. Further, since ORCA is a first-order trajectory simulator dependent only on relative configuration of neighbours, one can explain the performance of \texttt{D-Grid} being at par with \texttt{S-Grid} in the controlled setup.




\subsubsection{\textbf{Aggregation Strategy}} We focus on the information aggregation strategies for non-grid based encoders. It is evident that the baseline \texttt{D-MLP-Conc-MLP} of \textit{concatenating} the neighbourhood information performs better than the sophisticated attention-based \texttt{D-MLP-Attn-MLP} and max-pooling-based \texttt{D-MLP-MaxP-MLP} alternatives. This performance can be attributed to the simplicity of the concatenation scheme along with its property to preserve the identity of the surrounding neighbours. The MaxPooling strategy mixes up the different embeddings of the neighbours resulting in a high collision loss. 

\subsubsection{\textbf{LSTM-based interaction model}} Among the non-grid LSTM-based designs, the drop in performance of \texttt{D-MLP-SumPool-LSTM} module \cite{Ivanovic2018TheTP} can be attributed to (1) sum pooling which loses the individual identity of the neighbours and (2) encoding of absolute neighbour coordinates instead of relative coordinates: relational coordinates of agents to the target agent are easier to train than exact coordinates of agents. We notice that encoding the interaction information using LSTM [\texttt{O-LSTM-Att-MLP}, \texttt{D-MLP-Conc-LSTM}], improves performance over its MLP-based counterparts. MLP encoders, due to their non-recurrent nature, have no information regarding the interaction representation at the previous step. We argue that LSTMs can capture the evolution of interaction and therefore provide a better neighbourhood representation as the scene evolves, especially in cases where the input measurements are noisy. 

\subsection{Interaction Models: Real World Experiments}
Now, we discuss the performances of forecasting models on TrajNet++ real-world data. With the help of our defined trajectory categorization, we construct the \textit{TrajNet++} real-world benchmark by sampling trajectories corresponding mainly to \textit{Type III Interacting} category. Having gained insights on the performance of different modules on controlled synthetic data, we explore the question, `Do these findings generalize to the real world datasets comprising much more diverse interactions?'


\begin{table}[t]
\centering
\resizebox{0.48\textwidth}{!} {
\begin{tabular}{ |c|c|c|c| } 
 \hline
\textbf{Model (Acronym)} &  \textbf{ADE}/\textbf{FDE} & \textbf{Col-I} & \textbf{Col-II}\\ 
 \hline
 \hline
\multicolumn{4}{|c|}{\textbf{Hand-crafted methods}}\\
 \hline
 Kalman Filter                                  &  0.87/1.69 & 16.20  & 22.1\\
 \hline
 Social Force                                   &  0.89/1.53 & \textbf{0.0} & \textbf{13.1}\\
\hline 
 ORCA                                           & 0.68/1.40 & \textbf{0.0} & 15.0  \\
\hline
\hline
\multicolumn{4}{|c|}{\textbf{Top submitted methods*}\footnote{These methods are implemented by authors of the corresponding paper on TrajNet++}}\\
 \hline
 AMENet \cite{Cheng2020AMENetAM}                         & 0.62/1.30 & 14.1 & 16.90\\
 \hline
 AIN \cite{Zhu2020RobustTF}                              & 0.62/1.24 & 10.7 & 17.10\\
 \hline
 PecNet \cite{Mangalam2020ItIN}                          & 0.57/1.18 & 15.0 & 14.3\\

\hline
\hline
\multicolumn{4}{|c|}{\textbf{Grid based methods}}\\
\hline
 Vanilla                              & 0.6/1.3 & 13.6 (0.2) & 14.8 (0.1)\\ 
 \hline
O-Grid \cite{Alahi2016SocialLH}           & 0.58/1.24 & 9.1 (0.4)  & 15.1 (0.3)\\ 
\hline
S-Grid \cite{Alahi2016SocialLH}           & 0.53/1.14 & 6.7 (0.2) & 13.5 (0.5)\\
 \hline
D-Grid [\textbf{Ours}]                    & 0.56/1.22 & \textbf{5.4 (0.3)} & \textbf{13.0 (0.5)}\\
 \hline
 \hline
 \multicolumn{4}{|c|}{\textbf{Non-Grid based methods}}\\
\hline
  S-MLP-MaxP-MLP \cite{Gupta2018SocialGS}              & 0.57/1.24 & 12.6 (0.9) & 14.6 (0.7)\\
 \hline
 S-MLP-Att-MLP \cite{Kosaraju2019SocialBiGATMT}   & 0.56/1.22	& 7.2 (0.8)	& 14.8 (0.4)\\
 \hline
D-MLP-SumP-LSTM \cite{Ivanovic2018TheTP} & 0.60/1.28 & 13.9 (0.7) & 15.4 (0.5)\\
 \hline
O-MLP-Att-LSTM \cite{Vemula2017SocialAM}    & 0.56/1.21 & 9.0 (0.3) & 15.2 (0.4)\\
 \hline
D-MLP-ConC-MLP &       0.58/1.23 & 7.6 (0.6) & 14.3 (0.2)\\ 
 \hline
D-MLP-MaxP-MLP &       0.60/1.25 & 12.9 (0.6) & 14.8 (0.5)\\
 \hline
D-MLP-Attn-MLP &       0.56/1.22  & 6.9 (0.3) & 14.3 (0.6)\\
 \hline
D-MLP-ConC-LSTM (k=8) &       0.56/1.22 & 8.5 (0.5) & \textbf{14.0 (0.1)}\\ 
 \hline
 D-MLP-ConC-LSTM  [\textbf{Ours}]    &  0.55/1.19 & \textbf{6.8 (0.4)} & 14.5 (0.5)\\ 
 \hline
\end{tabular} }
\caption{Unimodal Comparison of interaction encoder designs when forecasting 12 future time-steps, given the previous 9 time-steps, on \textit{interacting} trajectories of TrajNet++ real world dataset. Errors reported are ADE / FDE in meters, Col I / Col II in  mean \% (std. dev. \%) across 5 independent runs. We emphasize that our goal is to reduce Col-I without compromising distance-based metrics.}
\label{Real_comparison}
\end{table}


Table~\ref{Real_comparison} provides an extensive evaluation of existing baselines on the Type III \textit{interacting} trajectories of the TrajNet++ real dataset. We observe that Col-I metric is the differentiating factor for various model designs when compared on \textit{identical grounds}. We hope that in future, researchers will incorporate the collision metrics while reporting their model performances on trajectory forecasting datasets. Moreover, the performance of ADE/FDE is similar ( including submitted methods) indicating that there exists a lot of scope to improve the performance of current trajectory forecasting models on a well-sampled interaction-centric test set.  


\subsubsection{\textbf{Classical Methods}} We first compare with the classical trajectory forecasting models, namely, Extended Kalman Filter (EKF), Social Force \cite{SocialForce}, and ORCA \cite{Berg2008ReciprocalVO}. Both Social Force and ORCA models forecast the future trajectory based on the assumption that each pedestrian has an intended direction of motion and a preferred velocity. We interpolate the observed trajectory to identify the \textit{virtual goals} for each agent. \texttt{Social Force} and \texttt{ORCA} are calibrated to fit the TrajNet++ training data by minimizing ADE/FDE metrics, along with the constraint that collisions should be avoided 

The high error of EKF can be attributed to the fact that the filter does not model social interactions. The interaction-based NN models outperform the handcrafted models in terms of the distance-based metrics, as NN have the ability to learn the subtle and diverse social interactions.

\subsubsection{\textbf{Grid-based modules}}  Our proposed \texttt{D-Grid} performs superior to \texttt{O-Grid} in the real world as well. It is interesting to compare the performances of \texttt{D-Grid} and \texttt{S-Grid}. The current design of \texttt{S-Grid} fails to learn the notion of prediction collision. This reaffirms the fact that while training to minimize ADE/FDE, the hidden-state of LSTM is unable to provide representations necessary to avoid collisions. In the \texttt{D-Grid} design, we force the model to focus explicitly on relative velocities based on our domain knowledge. The simplicity of our design slightly hampers the distance-based accuracy as we limit the expressibility of the model. However, it leads to safer predictions as the task of the model to learn social concepts is made easier thanks to our domain-knowledge based design. Further, as shown in Table~\ref{tab:compute}, the \texttt{D-Grid} provides significant computational speed-up in comparison to \texttt{S-Grid} rendering it useful for real-time deployment.


\begin{table}[h]
    \centering
    \begin{tabular}{|c|c|c|c|c|}
    \hline
                & Vanilla & O-Grid & S-Grid & D-Grid \\
    \hline
        Time    &     0.01    &   0.022  &  0.081  & 0.022 \\
    \hline
        Speed-Up &   8.1x & 3.7x & 1x & \textbf{3.7x}  \\  
    \hline
    \end{tabular}
    \caption{Speed (in seconds) comparison with S-Grid at test-time. D-Grid provides 3.7x speedup as compared to S-Grid rendering it more suitable for real-world deployment tasks.}
    \label{tab:compute}
\end{table}

\begin{figure*}
    \centering
    \includegraphics[width=0.98\textwidth]{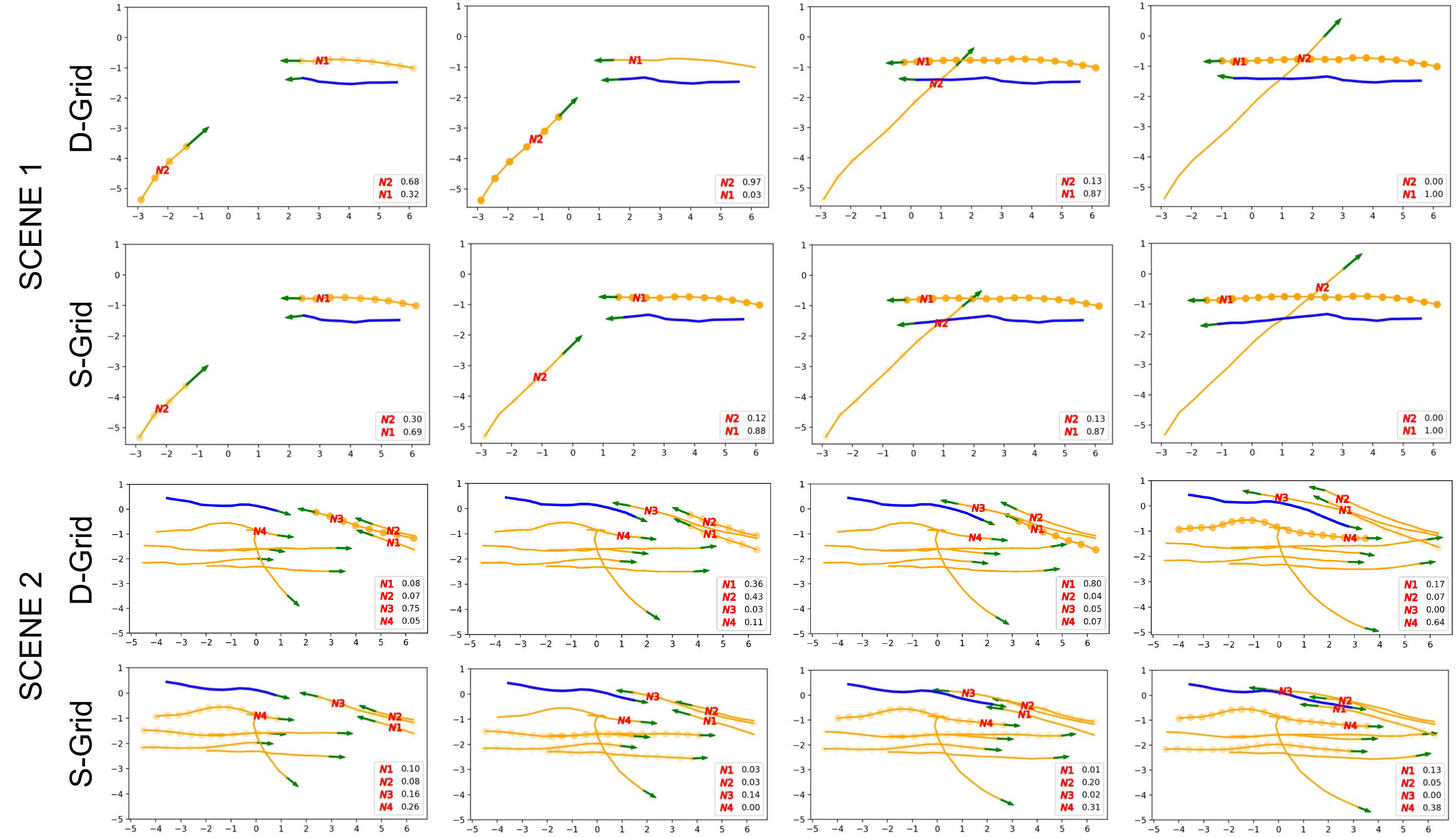}
    \caption{Visualizing the decision-making of grid-based interaction modules using layer-wise relevance propagation. The darker the yellow circles, the more is the weight (also shown in the legend) provided by the primary pedestrian (blue) to the corresponding neighbour (yellow). Our proposed \texttt{D-Grid}, driven by domain knowledge, outputs more human-like trajectories with more intuitive focus on surrounding neighbours as compared to \texttt{S-Grid}.}
    \label{fig:lrp_s19}
\end{figure*}

\subsubsection{\textbf{Aggregation Strategy}} 
We evaluate the performance of various aggregation strategies [\texttt{D-MLP-Attn-MLP}, \texttt{D-MLP-MaxP-MLP}, \texttt{D-MLP-ConC-MLP}] on real-world data keeping all the other factors constant. We observe that the max-pooling strategy performs the worst due to its design to hard-merge the embeddings of various neighbours. The concatenation strategy, despite its simplicity, performs only slightly worse in comparison to its sophisticated attention-based counterpart. One interesting point to note is that \texttt{D-MLP-Attn-MLP} performs superior to its social counterpart \texttt{S-MLP-Attn-MLP} further corroborating the strength of knowledge-based modules. We believe that the concatenation baseline is a simple yet powerful baseline to compare to when designing future information aggregating modules.

\subsubsection{\textbf{LSTM-based interaction models}} Among the LSTM-based non-grid designs, \texttt{D-MLP-SumPool-LSTM} module \cite{Ivanovic2018TheTP} demonstrates high Col-I metric due to (1) sum pooling strategy and (2) encoding of absolute neighbour coordinates. The Col-I metric for \texttt{O-LSTM-Att-MLP}  \cite{Vemula2017SocialAM} is relatively higher compared to \texttt{D-MLP-Concat-LSTM} in the real-world due to the absence of relative velocity as input to the interaction model. One can notice the importance of having an LSTM-based embedding in our proposed \texttt{DirectConcat} model by comparing the performance between \texttt{D-MLP-Concat-LSTM} and \texttt{D-MLP-Concat-MLP}. This design choice helps to model higher-order spatio-temporal interactions better and is more robust to noise in the real-world measurements as LSTM controls the evolution of the interaction vector. The top-$k$ neighbours are chosen based on euclidean distance. We argue that imposing domain knowledge by considering nearest neighbours is one of the reasons for improvement in Col-I metric as compared to its attention-based and max-pooling-based counterparts. This is corroborated by observing that considering a large number of nearest neighbours $(k=8)$, in comparison to $(k=4)$, results in an increase in the model prediction collisions. 

\subsubsection{\textbf{Comparison to Vanilla LSTM}}
The interaction-based models perform superior to \texttt{Vanilla LSTM} in terms of distance-based metrics. However, an important point to discuss is the performance comparison between \texttt{Vanilla LSTM} and interaction-based models in terms of the Col-II metric. We would like to remind that performance in Col-II metric represents the cases where the model predicts the correct mode for the primary pedestrian so that the collisions with the ground-truth trajectories of neighbours is minimal. Due to the multimodal nature of real-world data, it is quite possible that the interaction model predicts a different mode for one of the pedestrians (primary or neighbour) leading to the primary pedestrian not following the ground-truth mode. Indeed, two of the current interaction models [\texttt{O-MLP-Att-LSTM}, \texttt{D-MLP-SumP-LSTM}] struggle in accurately predicting the ground-truth mode compared to \texttt{Vanilla LSTM}. However, this observation \textit{does not} undermine the importance of modelling social interactions. The usefulness of modelling social interactions is justified by the Col-I metric comparison, which indicates that given the chosen mode for the primary pedestrian, the interaction models predicts a collision-free future for the entire scene, as opposed to \texttt{Vanilla LSTM}.




\subsubsection{Modified Training Objective} We employ a modified training objective where we penalize only the primary pedestrian in comparison to the standard practice of penalizing all pedestrians in the scene \cite{Gupta2018SocialGS, Vemula2017SocialAM, Ivanovic2018TheTP}. In the real world dataset, we know that the primary trajectories are largely interacting thanks to our defined categorization; however, there exist significant portion of trajectories among the neighbours which are static and linear. Penalizing such neighbouring trajectories during training might bias to the network into learning linear and static behavior because of the resulting imbalanced distribution (caused by the neighbours).

Table~\ref{train-ablation} illustrates the effectiveness of our modified training objective in helping the model to learn collision avoidance better. During test time, we do \textit{not} provide the ground truth neighbour trajectories.

\begin{table}[h]
\centering
\begin{center}
\resizebox{0.48\textwidth}{!}{\begin{tabular}{ |c|c|c|c|c| } 
 \hline
 Training Objective &  Dataset &   ADE & FDE & Col-I\\ 
 \hline
  Standard \cite{Gupta2018SocialGS, Vemula2017SocialAM, Ivanovic2018TheTP} & Synth  & 0.25 & 0.50 & 11.9 \\ 
 Proposed [\textbf{Ours}]  & Synth  & \textbf{0.24} & \textbf{0.49} & \textbf{2.2} \\  
\hline
 \hline
  Standard \cite{Gupta2018SocialGS, Vemula2017SocialAM, Ivanovic2018TheTP} & Real   & 0.59 & 1.27 & 7.4 \\  
 Proposed [\textbf{Ours}] & Real & \textbf{0.56} & \textbf{1.22} & \textbf{5.4}\\  
 \hline

\end{tabular}}
\end{center}
\caption{Our proposed training objective that penalizes only the primary prediction provides superior performance.}
\label{train-ablation}
\end{table}

\subsubsection{\textbf{Understanding NN decision-making}}
Now, using the popular technique of LRP, we investigate how various input factors affect the decision-making of the NN at each time-step. This helps us in verifying whether the NN decision-making process follows human intuition. Fig~\ref{fig:lrp_s19} illustrates the score of each neighbour obtained on applying the LRP procedure on our proposed \texttt{D-Grid} module and baseline \texttt{S-Grid} in real-world scenarios.

In Scene 1, we demonstrate the application of LRP on a simple real-world example. In case of \texttt{D-Grid}, the primary pedestrian starts focusing on the potential collider $N2$ despite it being distant compared to $N1$ thereby preventing collision by staying closer to $N1$. On the other hand, \texttt{S-Grid} keeps focusing on the $N1$ which is not desirable. It is interesting to note that once $N2$ passes the primary pedestrian, both \texttt{D-Grid} and \texttt{S-Grid} shift the attention of the primary pedestrian back to $N1$.

In Scene 2, we demonstrate the effectiveness of our proposed \texttt{D-Grid} module in a complex real-world scenario. For \texttt{D-Grid}, initially the primary pedestrian focuses on $N3$ to prevent collision. On successfully avoiding collision with $N3$, \texttt{D-Grid} immediately shifts the focus to the pair $N1$ and $N2$ as they would potentially lead to another collision. On coming in close proximity to $N1$ and $N2$, the focus significantly shifts towards  $N1$ as it is closer to the primary pedestrian. Finally, on passing $N1$ and $N2$, the primary pedestrian attends to the pedestrian $N4$ in front. On the other hand, \texttt{S-Grid} passes in between $N1$ and $N2$, such behavior is not expected in human crowds.

Thus, we can see that LRP is an effective investigative tool to understand the rationale behind the NN decisions. We can observe that, along with having a lower Col-I metric as compared to \texttt{S-Grid} in Table~\ref{Real_comparison}, the decision making of our domain-knowledge based \texttt{D-Grid} satisfies human intuition while navigating crowds. The LRP technique is generic and can be applied on top of any existing trained interaction module architecture.

To summarize, despite claims in literature that specific interaction modules better model interactions, we observe that under \textit{identical} conditions, all modules perform similar in terms of the distance-based ADE and FDE metrics. The incorporation of Col-I metrics paints a more complete picture of model performance.
Secondly, relative velocity plays a crucial role in learning collision avoidance in the real-world. Thirdly, a simple concatenation strategy performs at par with the sophisticated attention-based counterparts. We believe that the concatenation baseline should be a standard baseline to compare to when designing future information aggregating modules. Finally, the LRP technique is a useful investigative tool to gain insights regarding the decision-making process of NNs. We hope that such practices will help to accelerate the development of interaction modules in future research. There certainly exists room for improvement, and we hope that our benchmark provides the necessary resources to advance the field of trajectory forecasting. We open-source our code for reproducibility. 

\section{Conclusions}
In this work, we tackled the challenge of modelling social interactions between pedestrians in crowds. While modelling social interactions is a central issue in human trajectory forecasting, the literature lacks a definitive comparison between the many existing interaction model designs on identical grounds. We presented an in-depth analysis of the design of interaction modules proposed in the literature and proposed two domain-knowledge based interaction models.

A significant yet missing component in this field is an objective and informative evaluation of these interaction-based methods. To solve this issue, we propose \textit{TrajNet++}: (1) TrajNet++ is interaction-centric as it largely comprises scenes where interactions take place thanks to our defined trajectory categorization, both in the real-world and synthetic settings, (2) TrajNet++ provides an extensive evaluation system that includes novel collision-based metrics that can help measure the \textit{physical feasibility} of model predictions. The superior quality of TrajNet++ is highlighted by the improved performance of interaction-based models on real world datasets on all metrics (4 of the top 5 methods on TrajNet \cite{sadeghiankosaraju2018trajnet}, an earlier benchmark, do not model social interactions). Further, we demonstrated how our collision-based metrics provide a more concrete picture regarding the model performance. 

Our proposed models outperform competitive baselines on TrajNet++ synthetic dataset by benchmarking against several popular interaction module designs in the field. On the real dataset, there is no clear winner amongst all the designs in terms of distance-based metrics, when compared on equal grounds. Our proposed designs show significant gains in reducing model prediction collisions. There is room for improvement, and we hope that our benchmark facilitates researchers to objectively and easily compare their methods against existing works so that the quality of trajectory forecasting models can keep increasing, allowing us to tackle more challenging scenarios.

\newpage
\ifCLASSOPTIONcaptionsoff
  \newpage
\fi

\bibliographystyle{IEEEtran}
\bibliography{newstuff.bib}{}

\newpage

\begin{center}
\textbf{\large Human Trajectory Forecasting in Crowds:} \\
\textbf{\large A Deep Learning Perspective} \\
\vspace{1em}
\textbf{\large Supplementary Material} \\
\end{center}

\section{Trajnet++ Framework}

\subsection{Categorization Rules}
In this section, we provide the mathematical conditions to be satisfied to classify a scene into a particular category. 

\begin{itemize}
\item \textbf{Static (Type I)}: The total distance traveled by the primary pedestrian is \textit{less than 1 meter}.
\item \textbf{Linear (Type II)}: A trajectory is categorized to be \textit{Linear} if the final displacement error (FDE) of the \textit{extended kalman filter prediction is less than 0.5m}. 
\item \textbf{Leader Follower (Type IIIa)}: The primary pedestrian follows a neighbour, moving in the same direction within an angular range of $\pm 15 \deg$ and having relative velocity in the range of $\pm 15 \deg$, for more than 2 seconds.
\item \textbf{Collision Avoidance (Type IIIb)}: The primary pedestrian is faced head-on by a neighbour, coming from the opposite direction within an angular range of $\pm 15 \deg$ and relative velocity of movement in the range [$180 \pm 15 \deg$]) at any given point of time.
\item \textbf{Group (Type IIIc)}: There exists a neighbour on the side of the primary pedestrian (angular range [$90 \pm 15 \deg$] or [$ -90 \pm 15 \deg$]) for the entire duration of the sample with mean distance $\leq 1 m$ and standard deviation $\leq 0.2 m$.
\item \textbf{Other Interactions (Type IIId)}: If at any point of time, a neighbour exists in the front (angular range of $\pm 15 \deg$) of the primary pedestrian within a distance of 5 m, we categorize this trajectory of undergoing social interactions.
\item \textbf{Non-Interacting (Type IV)}: All the other trajectories that fail to satisfy any of the above conditions.
\end{itemize}

\subsection{TrajNet++ Datasets}
We now describe the datasets used in the TrajNet++ benchmark. Since the focus of this work is to tackle agent-agent interactions in crowded settings, we explicitly select datasets where scene constraints do not play a significant role in determining the future trajectory. For each real world dataset, we utilize only the information regarding the pedestrian locations from the respective annotations files, \textit{i.e.}, spatial coordinates of each pedestrian at each time frame. Furthermore, we provide no information regarding the destination of each pedestrian or structure of the scene. Our goal is to forecast only the 2D spatial coordinates for each pedestrian.

\subsubsection{TrajNet++ Real Dataset}
\begin{itemize}
\item \textbf{ETH:} ETH dataset provides for two locations: Univ and Hotel, where pedestrian trajectories are observed. This dataset contains a total of approximately 750 pedestrians exhibiting complex interactions (Pellegrini \textit{et. al.} \cite{Pellegrini2010ImprovingDA}). The dataset is one of the widely used benchmarks for pedestrian trajectory forecasting. It captures diverse real-world social interactions like leader follower, collision avoidance, and group forming and dispersing.
\item \textbf{UCY:} UCY dataset consists of three scenes: Zara01, Zara02 and Uni, with a total of approximately 780 pedestrians (Lerner \textit{et. al.} \cite{Lerner2007CrowdsBE}). This dataset, in addition to the ETH dataset, is widely used as benchmarks for pedestrian trajectory forecasting, offering a wide range of non-linear trajectories arising out of social interactions.
\item \textbf{WildTrack:} This is a recently proposed benchmark \cite{Chavdarova2018WILDTRACKAM} for pedestrian detection and tracking captured in front of ETH Zurich. Since the dataset comprises of diverse crowd interactions in the wild, we utilize it for our task of trajectory forecasting. 
\item \textbf{L-CAS:}  This is a recently proposed benchmark for pedestrian trajectory forecasting (Sun \textit{et. al.} \cite{Sun20173DOFPT}). The dataset, comprising over 900 pedestrian tracks, comprises diverse social interactions that are captured within indoor environments. Some of the challenges scenarios in this dataset include people pushing trolleys and running children.
\item \textbf{CFF:} This is a large-scale dataset of 42 million trajectories extracted from real-world train stations \cite{Alahi2014SociallyAwareLC}. It is one of the biggest datasets that capture agent-agent interactions in crowded settings during peak travel times. Due to the high density of people, we observe higher instances of social interactions like leader-follower in this dataset. 
\end{itemize}

\subsubsection{TrajNet++ Synthetic Dataset}\label{sec:synth_data}
Interaction-centric synthetic datasets can provide the necessary controlled environment to compare the performances of different model components. We provide synthetic data in TrajNet++ to evaluate the performance of a model under controlled interaction scenarios. 

\begin{figure}
\centering
    \includegraphics[width=0.45\textwidth]{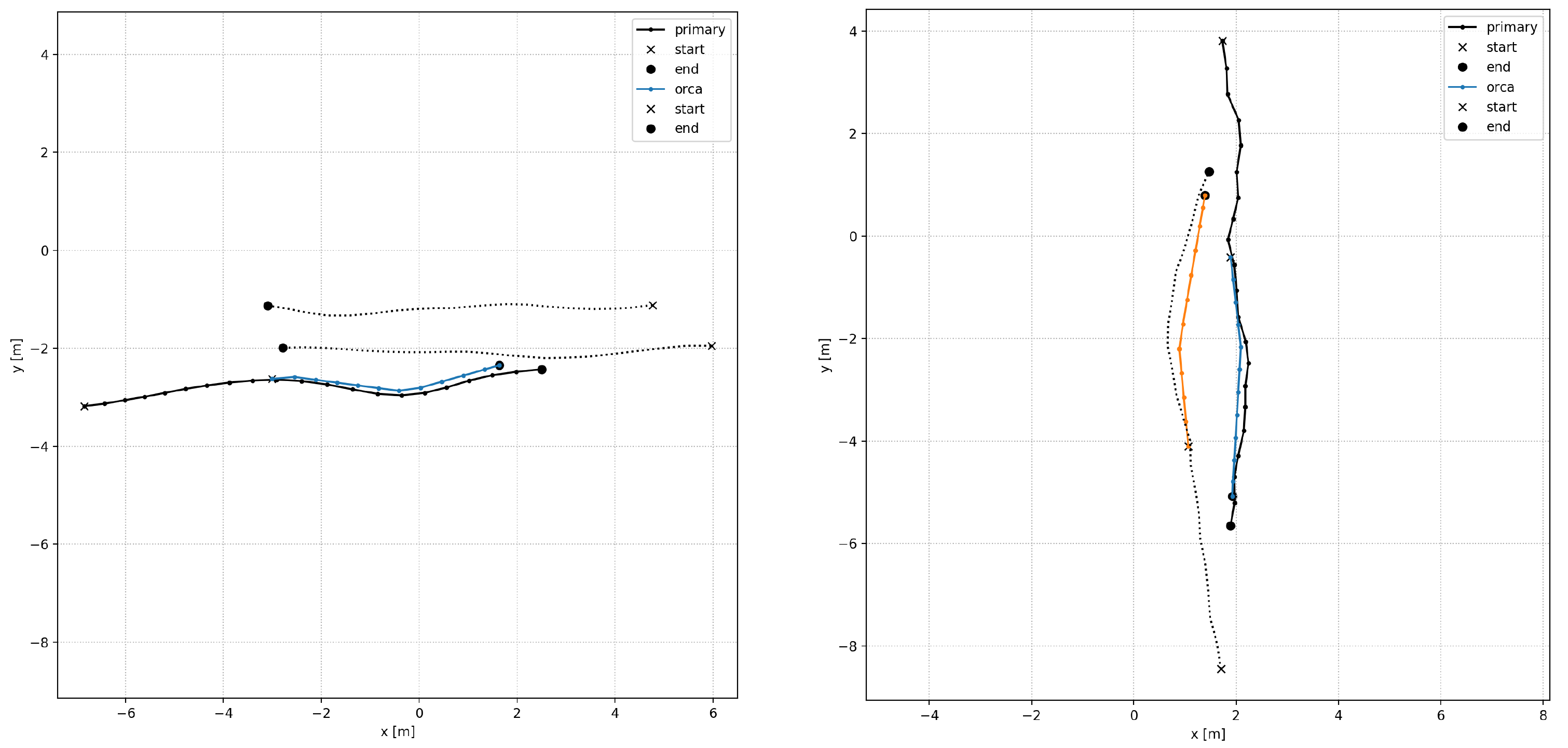}
    \caption{Illustration of our calibrated ORCA parameters showing similar reaction curvature (in blue) as those shown by humans in real world datasets (in black). Solid line denotes the primary pedestrian. Dotted lines denote the neighbours.}
    \label{fig:calibration}
\end{figure}

\textbf{Simulator Selection:} It is a necessary condition that the interactions in the synthetic dataset are similar to those in the real world. Empirically, we find that in comparison to Social Force \cite{SocialForce}, ORCA \cite{Berg2008ReciprocalVO} provides a better similarity to real world human motion with respect to collision avoidance. We choose ORCA parameters, which demonstrate a reaction distance and reaction curvature similar to real data during collision avoidance (Fig~\ref{fig:calibration}).

\textbf{Dataset Generation:} Given the ORCA parameters, we generated the synthetic dataset using the following procedure: $n$ pedestrians were initialized at random on a circle of radius $r$ keeping a certain minimum distance $d\_min$ between their initial positions. The goal of each pedestrian was defined to be the point diametrically opposite to the initial position on the circle. For the TrajNet++ synthetic dataset: We ran different simulations with $n$ chosen randomly from the range [4, 7) on a circle of radius $r = 10$ meters and $d\_min = 2$ meters. 

\begin{figure}[h]
\begin{subfigure}[h]{0.48\textwidth}
\includegraphics[width=0.48\textwidth]{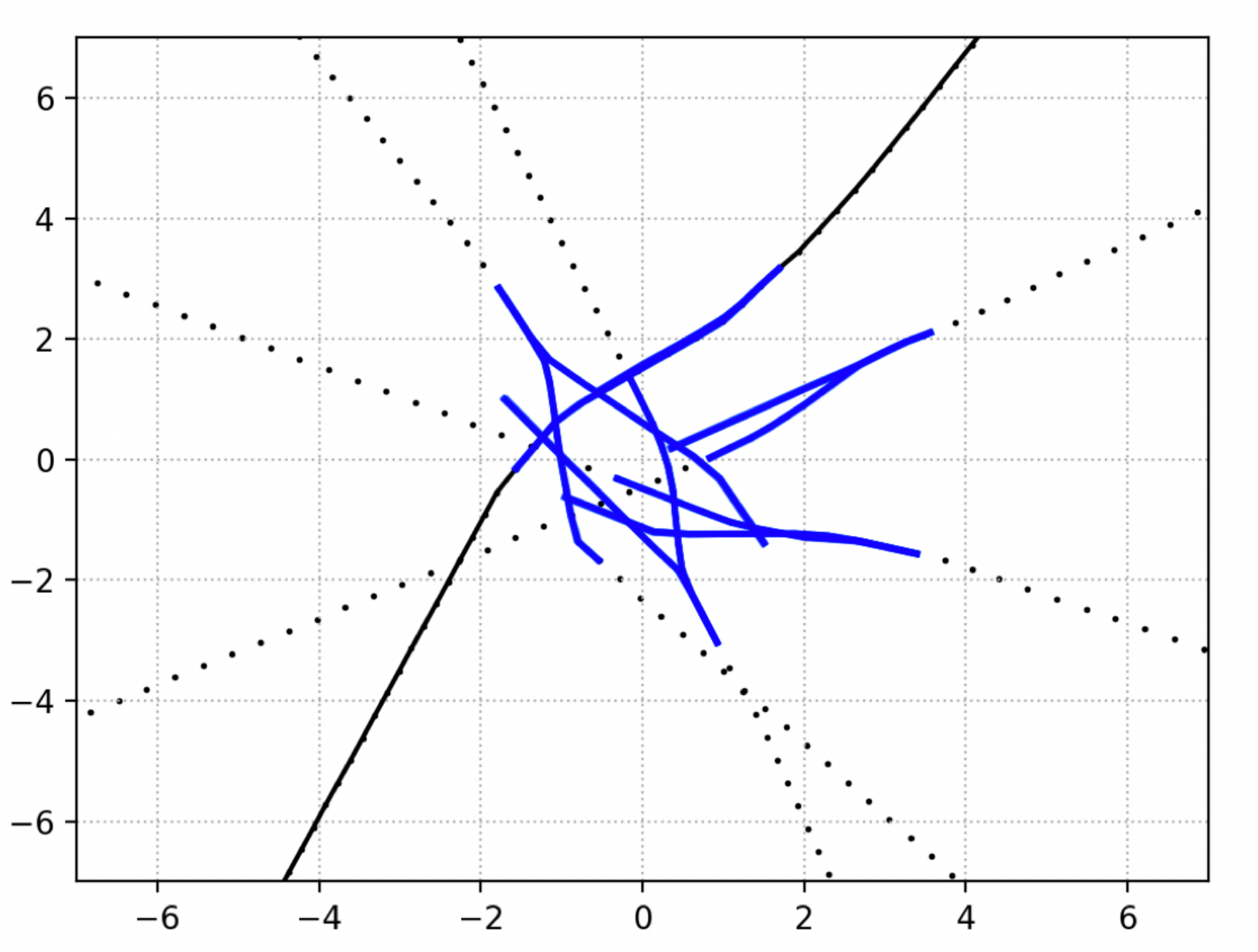}
\includegraphics[width=0.48\textwidth]{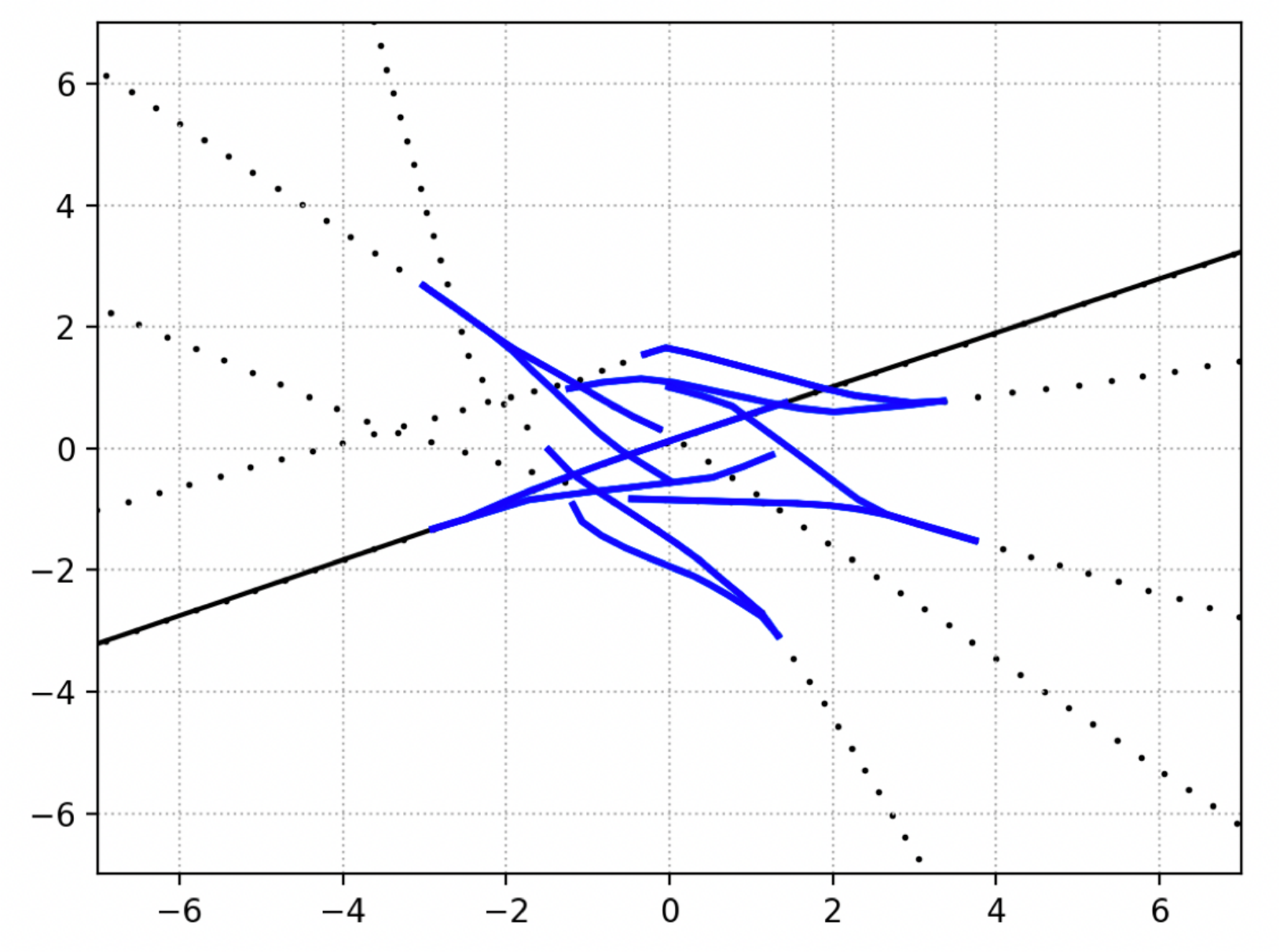}
\caption{Sensitive Scenes}
\end{subfigure}
\\
\begin{subfigure}[h]{0.48\textwidth}
\includegraphics[width=0.48\textwidth]{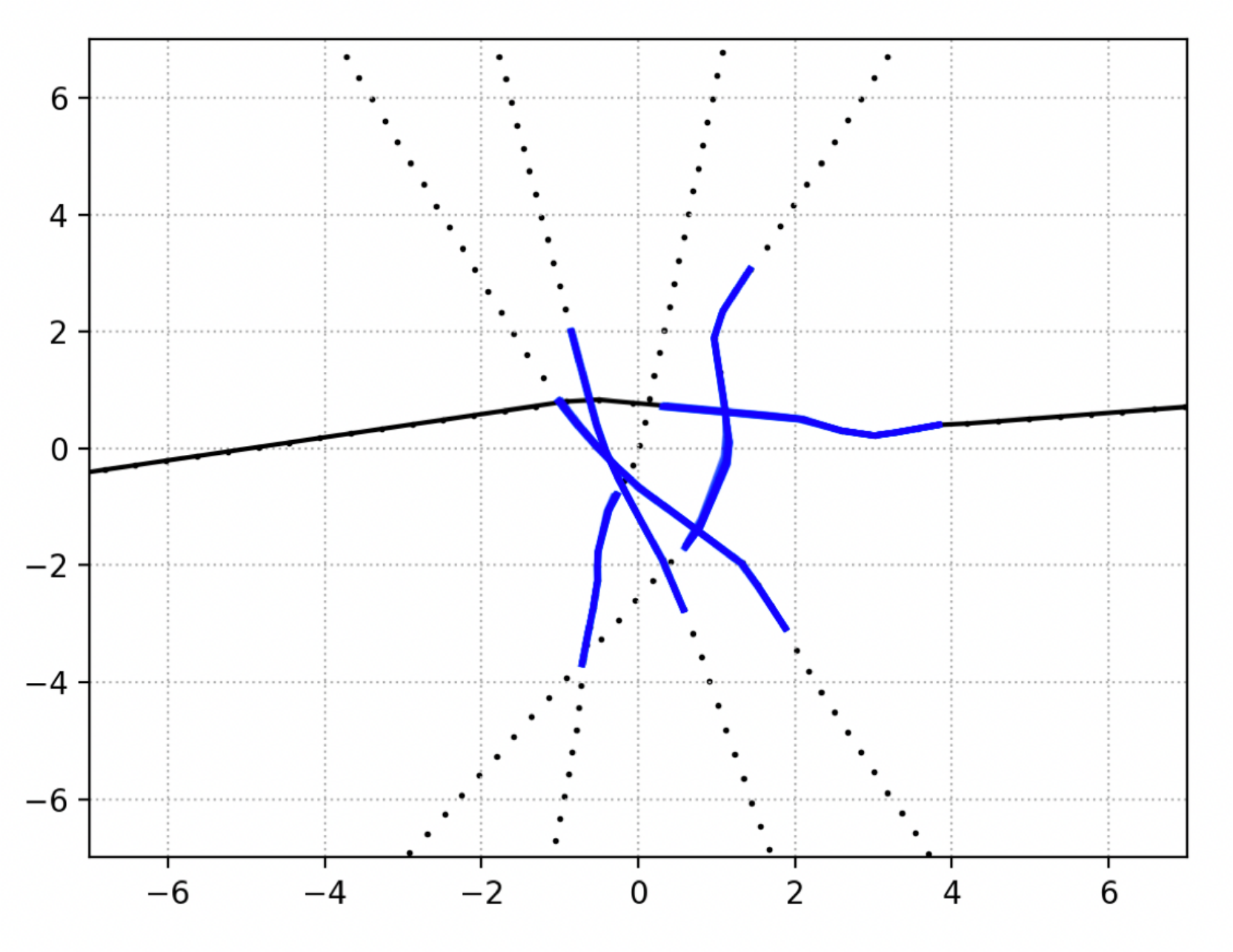}
\includegraphics[width=0.48\textwidth]{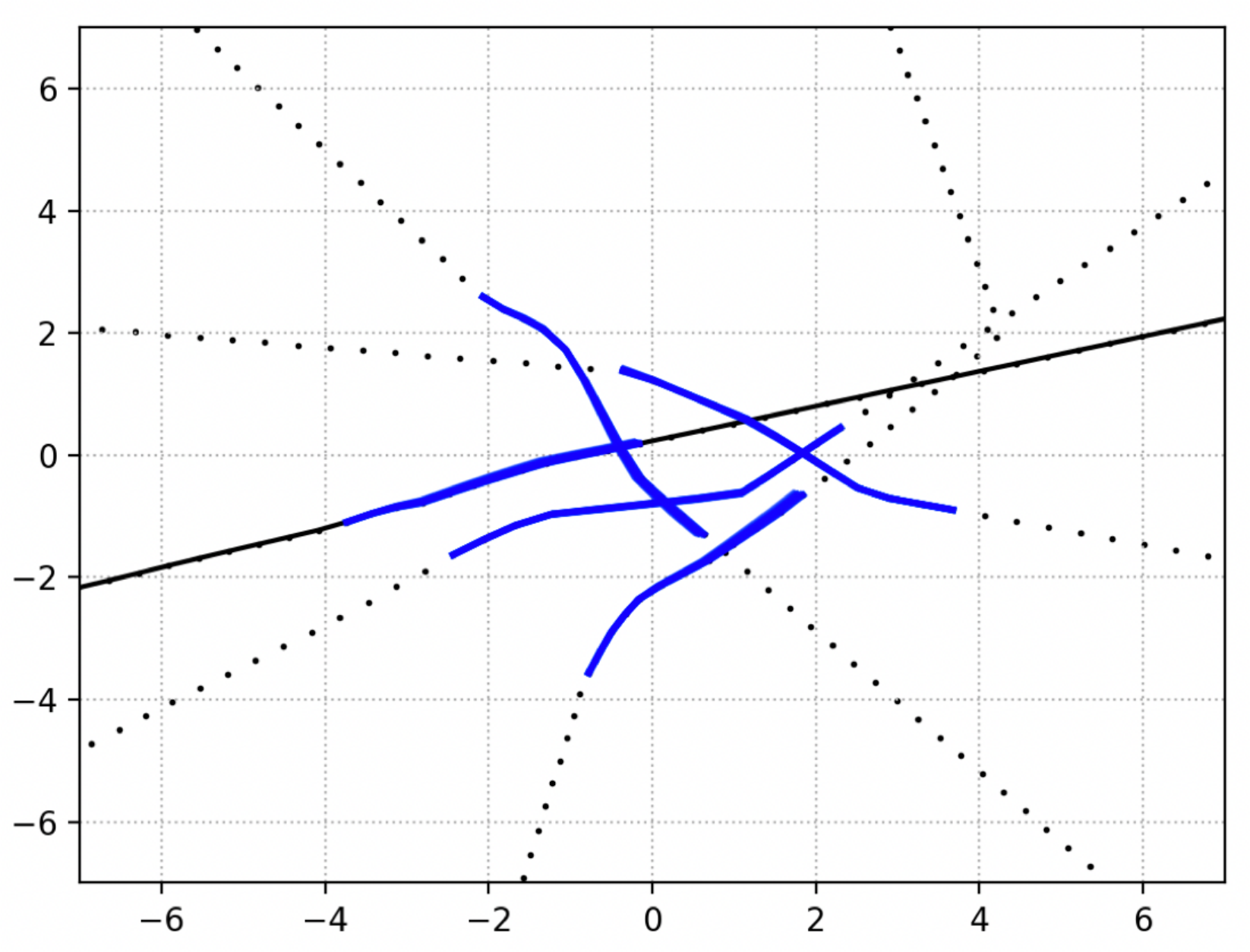}
\caption{Insensitive Scenes}
\end{subfigure}
\caption{Illustration of our filtering procedure to generate Trajnet++ synthetic dataset. Given a ground-truth scene (in black) generated by ORCA, we perturb the positions of agents and forecast the future with ORCA, iteratively, to obtain a distribution (in blue). This procedure helps us identify the \textit{sensitive scenes} and consequently remove them.}
\label{fig:sensitive}
\end{figure}

\begin{figure}
\centering
\begin{subfigure}[h]{0.15\textwidth}
    \includegraphics[width=\textwidth]{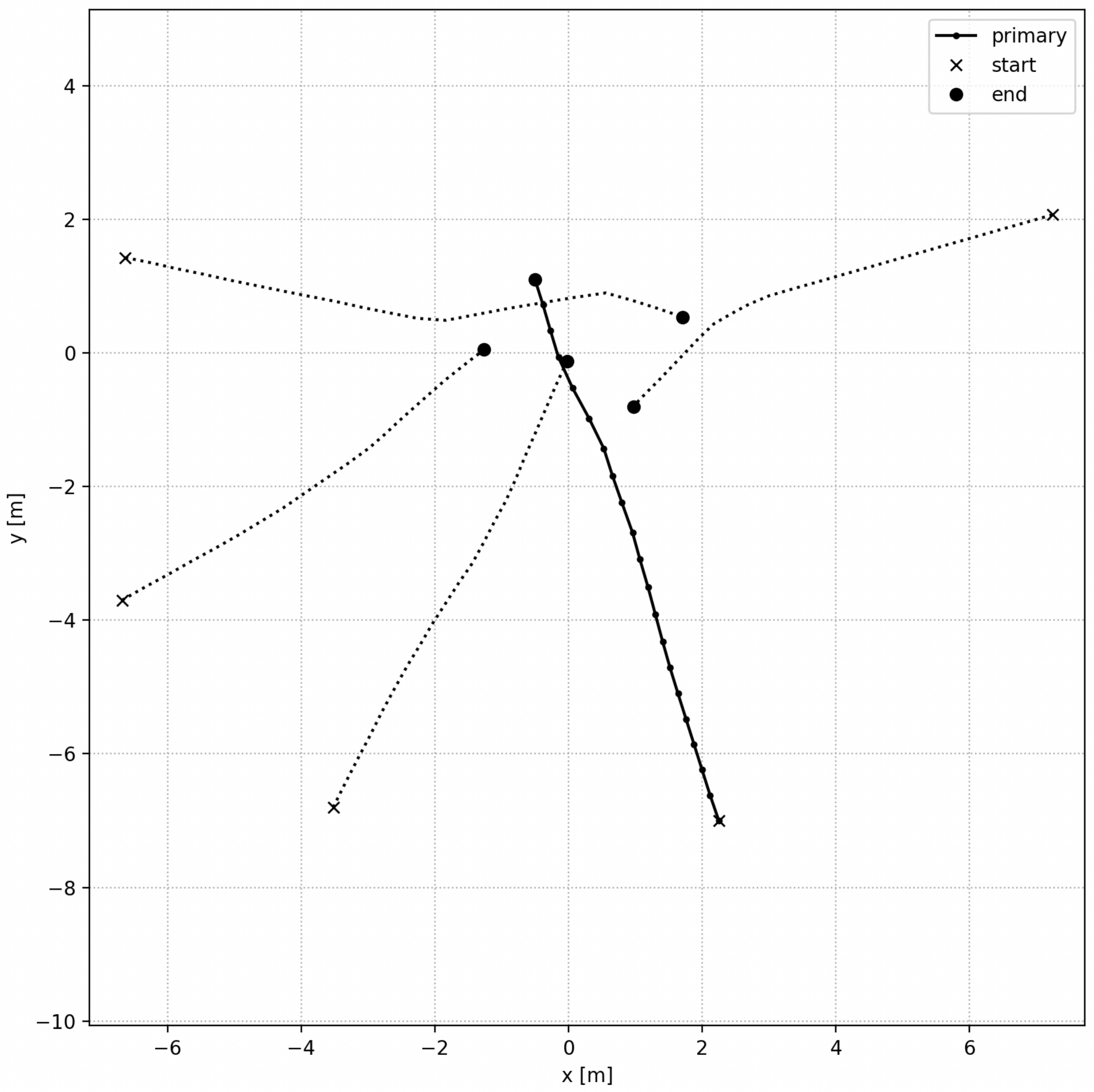}
\end{subfigure}
\begin{subfigure}[h]{0.15\textwidth}
    \includegraphics[width=\textwidth]{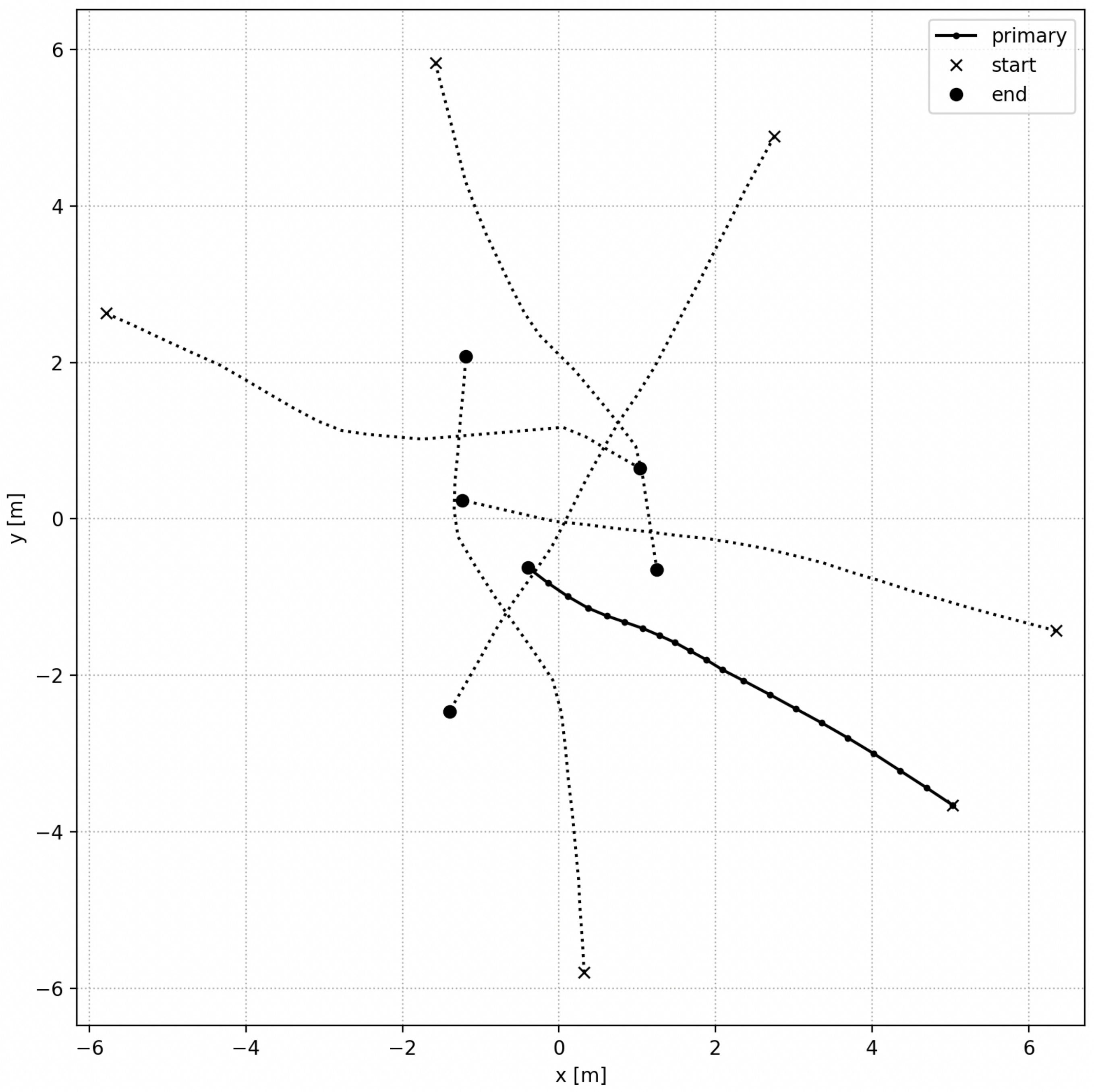}
\end{subfigure}
\begin{subfigure}[h]{0.15\textwidth}
    \includegraphics[width=\textwidth]{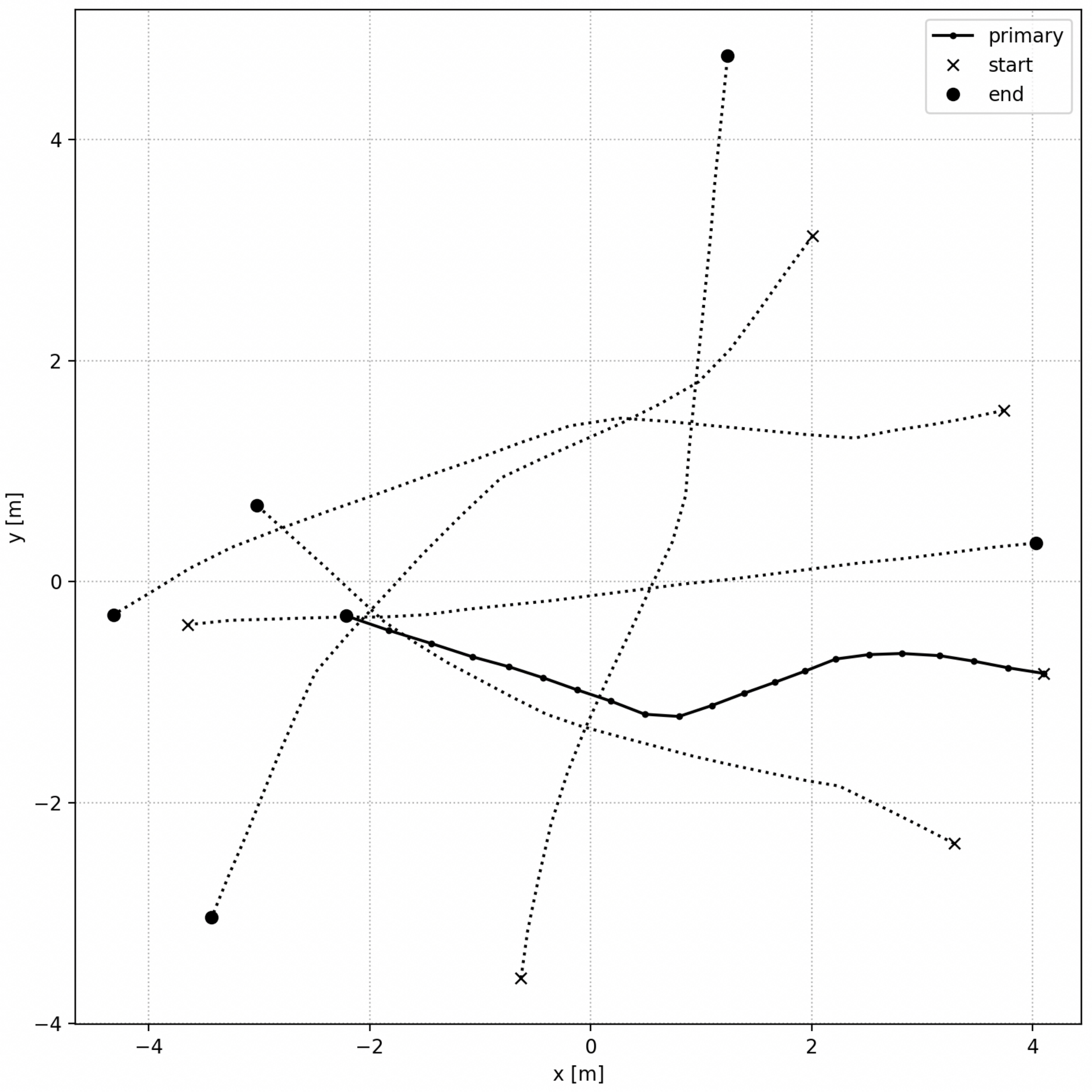}
\end{subfigure}
    \caption{Illustration of our synthetically generated samples using the calibrated ORCA parameters.}
    \label{fig:synth_traj}
\end{figure}

Given the generated trajectories, we selected only those scenes which belonged to the Type III: `Interacting' category. The ORCA simulator demonstrates sensitive dependence on initial conditions. 
This can be attributed to the fact that all the agents are expected to collide near the same point (at the origin), so slight perturbations can greatly affect the future trajectory of all agents.
Sensitivity to initial conditions, also known as the \textit{Butterfly Effect}, is a well-studied phenomenon of \textit{Chaos theory}\cite{Edward1972DoesTF}. To identify such sensitive initial conditions, the practice which is often followed is to perturb the initial conditions with arbitrary small noise and observe the effect. Along similar lines, we propose an additional step to filter out such `sensitive' scenes: in each scene, we perturb all trajectories at the point of observation with a small uniform noise ($noise \in U[-noise\_thresh, noise\_thresh]$), and forecast the future trajectories using ORCA. We perform this procedure $k$ times. If any of the $k$ ORCA predictions have a significant ADE compared to the ground truth, we filter out such scenes. Fig~\ref{fig:sensitive} visualizes the sample outputs of our filtering process (with $noise\_thresh = 0.01$, $k=20$, $n=5$). We passed the selected scenes through a final additional filter that identifies sharp unrealistic turns in trajectories. 
Fig~\ref{fig:synth_traj} illustrates a few sample scenes in our TrajNet++ synthetic dataset. 

\subsubsection{TrajNet++ Dataset Split}
We now provide the training dataset and test dataset split of the TrajNet++ benchmark (see Table~\ref{Train} and Table~\ref{Test}). The trajectories are divided according to our defined categorization for TrajNet++. 

\begin{center}
\begin{table}[htb]
\centering
\resizebox{0.48\textwidth}{!}{\begin{tabular}{ |c|c|c|c|c|c|c|c|c|c| } 
\hline
Name            &   Total    &  I  &  II & III & LF  &  CA  & Grp  & Oth  & IV \\ 
\hline
Synthetic       & 54513    &  0  &  0 & 54513 & 495 & 7183 & 0 & 46853  &  0  \\
\hline
BIWI Hotel.     & 229    & 13  &  91 & 109 &  22  & 29   &  38  &  41  & 16\\
Zara01.         & 1017   &  4 & 184 & 541 &  111 & 160  &  231 &  132  & 288\\
Zara03.         & 955    &  11 & 151 & 636 &  106  & 222  &  236 &  200 & 157\\
Stud01.         & 5265   & 102  & 564 & 4349& 651  & 1824 & 1665 & 1258 & 250\\
Stud03.         & 4262   &  39 & 342 & 3579& 535  & 1488  & 1455 & 1123 & 302\\
WildTr.         & 1101   & 119 &  43 & 667 &  41  & 76   &  144 &  423 & 272\\
L-CAS           & 889    & 215 &  87 & 297 &  10  & 86   &  6  &  202 & 292\\

\hline
CFF06         & 23751 & 105 & 3722 & 18980 &  6734 &  10950 & 323 & 5394 & 944\\
CFF07         & 24100 & 93 & 3685 & 19389 &  7139 &  11118 & 297 & 5410 & 933\\
CFF08         & 23070 & 106 & 3512 & 18456 &  6434 &  10438 & 309 & 5315 & 996\\
CFF09         & 11433 & 54 & 2260 & 8129  &  2284 &  3965 & 236 & 3006 & 990\\
CFF10         & 9702  & 47 & 2028 & 6825  &  1890 &  3397 & 181 & 2499 & 802\\
CFF12         & 23936 & 87 & 3740 & 19141 &  6948 &  11013 & 326 & 5320 & 968\\
CFF13         & 22355 & 102 & 3348 & 17906 &  6112 &  9913 & 300 & 5352 & 999\\
CFF14         & 23376 & 90 & 3566 & 18693 &  6697 &  10623 & 318 & 5364 & 1027\\
CFF15         & 22657 & 106 & 3354 & 18201 &  6218 &  10238 & 306 & 5362 & 996\\
CFF16         & 10771 &  61 & 2116 &  7619 &  2137 &  3564 & 250 & 2844 & 975\\
CFF17         & 10000 &  68 & 2204 &  6892 & 1917 &  3557 & 223 & 2408 & 836\\
CFF18         & 22021 & 88 & 3553 & 17317 &  6145 &  9644 & 312 & 5082 & 1063\\
\hline
\textbf{Total}& \textbf{250k} & & & & & & & &\\
\hline
\end{tabular}}
\caption{TrajNet++: Statistics of the Training Split.}
\label{Train}
\end{table}
\end{center}

\begin{center}
\begin{table}[htb]
\centering
\resizebox{0.48\textwidth}{!}{\begin{tabular}{ |c|c|c|c|c|c|c|c|c|c| } 
\hline
Name            &   Total    &  I  &  II & III & LF  &  CA  & Grp  & Oth  & IV \\ 
\hline
Synthetic        &  3842  & 0   &  0  &3842& 73  & 632 & 0    & 3142 & 0 \\
\hline
BIWI ETH.       &   1137   &  12  & 226 & 639 & 192  & 152  & 172  & 244  & 260 \\
UNI.            &   243    &  1  & 50  & 99 &   7  &  11  &  39  &  47  &  93 \\
Zara02.         &   1766   & 91  & 439 & 938 & 185  & 317  & 434  & 260  & 298 \\
\hline
\textbf{Total}  & \textbf{6988} & & & & & & & &  \\
\hline
\end{tabular}}
\caption{TrajNet++: Statistics of the Testing Split.}
\label{Test}
\end{table}
\end{center}

\end{document}